\documentclass[3p,preprint,12pt]{elsarticle}
\makeatletter\if@twocolumn\PassOptionsToPackage{switch}{lineno}\else\fi\makeatother

\usepackage[section]{placeins}
\usepackage{flafter}  
\usepackage{tabulary,xcolor}
\usepackage{amsfonts,amsmath,amssymb}
\usepackage[T1]{fontenc}
\usepackage{moreverb}
\usepackage{multicol}
\usepackage{amsmath}
\usepackage{relsize}
\usepackage{url,multirow,morefloats,floatflt,cancel,tfrupee}

\usepackage{algorithm}
\usepackage[noend]{algpseudocode}
\usepackage{hyperref}
\usepackage{systeme}

\usepackage{dsfont}
\usepackage{algorithm}
\usepackage{algorithmicx}
\usepackage{algpseudocode}
\usepackage{mathtools}
\newcommand{\minbox}[2]{%
  \mathmakebox[\ifdim#1<\width\width\else#1\fi]{#2}}

\algnewcommand{\Local}{\State\textbf{local variables: }}

\makeatletter
\let\save@ps@pprintTitle\ps@pprintTitle
\def\hlinewd#1{%
	\noalign{\ifnum0=`}\fi\hrule \@height #1%
	\futurelet\reserved@a\@xhline}

\AtBeginDocument{\ifNAT@numbers \biboptions{sort&compress}\fi}
\makeatother

\usepackage{etoolbox}
\makeatletter
\patchcmd{\hdots@for}{\hfill}{\hskip\z@\@plus 1filll}{}{}
\makeatother  

\usepackage{ifluatex}
\ifluatex
\usepackage{fontspec}
\defaultfontfeatures{Ligatures=TeX}
\usepackage[]{unicode-math}
\unimathsetup{math-style=TeX}
\else 
\usepackage[utf8]{inputenc}
\fi 
\ifluatex\else\usepackage{stmaryrd}\fi

\usepackage{url,multirow,morefloats,floatflt,cancel,tfrupee}
\makeatletter

\usepackage{indentfirst}
\AtBeginDocument{\@ifpackageloaded{textcomp}{}{\usepackage{textcomp}}}
\makeatother
\algnewcommand\algorithmicswitch{\textbf{switch}}
\algnewcommand\algorithmiccase{\textbf{case}}
\algnewcommand\algorithmicassert{\texttt{assert}}
\algnewcommand\Assert[1]{\State \algorithmicassert(#1)}%
\algdef{SE}[SWITCH]{Switch}{EndSwitch}[1]{\algorithmicswitch\ #1\ \algorithmicdo}{\algorithmicend\ \algorithmicswitch}%
\algdef{SE}[CASE]{Case}{EndCase}[1]{\algorithmiccase\ #1}{\algorithmicend\ \algorithmiccase}%
\algtext*{EndSwitch}%
\algtext*{EndCase}%
\usepackage{colortbl}
\usepackage{xcolor}
\usepackage{pifont}
\usepackage[nointegrals]{wasysym}
\usepackage{amssymb}

\let\emptyset\varnothing
\makeatletter

\def\mcWidth#1{\csname TY@F#1\endcsname+\tabcolsep}

\def\cAlignHack{\rightskip\@flushglue\leftskip\@flushglue\parindent\z@\parfillskip\z@skip}
\def\rAlignHack{\rightskip\z@skip\leftskip\@flushglue \parindent\z@\parfillskip\z@skip}

\if@twocolumn\usepackage{dblfloatfix}\fi 
\AtBeginDocument{
	\expandafter\ifx\csname eqalign\endcsname\relax
	\def\eqalign#1{\null\vcenter{\def\\{\cr}\openup\jot\m@th
			\ialign{\strut$\displaystyle{##}$\hfil&$\displaystyle{{}##}$\hfil
				\crcr#1\crcr}}\,}
	\fi
}

\AtBeginDocument{%
	\@ifpackageloaded{endfloat}%
	{\renewcommand\efloat@iwrite[1]{\immediate\expandafter\protected@write\csname efloat@post#1\endcsname{}}}{}%
}%

\let\lt=<
\let\gt=>
\def\processVert{\ifmmode|\else\textbar\fi}

\@ifundefined{subparagraph}{
	\def\subparagraph{\@startsection{paragraph}{5}{2\parindent}{0ex plus 0.1ex minus 0.1ex}%
		{0ex}{\normalfont\small\itshape}}%
}{}

\newcommand\role[1]{\unskip}
\newcommand\aucollab[1]{\unskip}

\@ifundefined{tsGraphicsScaleX}{\gdef\tsGraphicsScaleX{1}}{}
\@ifundefined{tsGraphicsScaleY}{\gdef\tsGraphicsScaleY{.9}}{}
\def\checkGraphicsWidth{\ifdim\Gin@nat@width>\linewidth
	\tsGraphicsScaleX\linewidth\else\Gin@nat@width\fi}

\def\checkGraphicsHeight{\ifdim\Gin@nat@height>.9\textheight
	\tsGraphicsScaleY\textheight\else\Gin@nat@height\fi}

\def\fixFloatSize#1{}%\@ifundefined{processdelayedfloats}{\setbox0=\hbox{\includegraphics{#1}}\ifnum\wd0<\columnwidth\relax\renewenvironment{figure*}{\begin{figure}}{\end{figure}}\fi}{}}
\let\ts@includegraphics\includegraphics

\def\inlinegraphic[#1]#2{{\edef\@tempa{#1}\edef\baseline@shift{\ifx\@tempa\@empty0\else#1\fi}\edef\tempZ{\the\numexpr(\numexpr(\baseline@shift*\f@size/100))}\protect\raisebox{\tempZ pt}{\ts@includegraphics{#2}}}}

\AtBeginDocument{\def\includegraphics{\@ifnextchar[{\ts@includegraphics}{\ts@includegraphics[width=\checkGraphicsWidth,height=\checkGraphicsHeight,keepaspectratio]}}}

\def\URL#1#2{\@ifundefined{href}{#2}{\href{#1}{#2}}}

\makeatother

\usepackage{graphicx}
\usepackage{subcaption}
\graphicspath{ {Figures/} }
\begin{document}
	\begin{frontmatter}
		
		\title{Parsimony-Enhanced Sparse Bayesian Learning for Robust Discovery of Partial Differential Equations}
		
		\author{Zhiming Zhang}
		\ead{zzhan506@asu.edu}  
		\author{Yongming Liu}
		\ead{corresponding author:yongming.liu@asu.edu}
		\address {School for Engineering of Matter, Transport and Energy, Arizona State University, Tempe, AZ, 85281, USA} 

% Please include an abstract:
\begin{abstract}
	Robust physics discovery is of great interest for many scientific and engineering fields. Inspired by the principle that a representative model is the one simplest possible, a new model selection criteria considering both model's Parsimony and Sparsity is proposed. A Parsimony-Enhanced Sparse Bayesian Learning (\textsc{Pe}SBL) method is developed for discovering the governing Partial Differential Equations (PDEs) of nonlinear dynamical systems. Compared with the conventional Sparse Bayesian Learning (SBL) method, the \textsc{Pe}SBL method promotes parsimony of the learned model in addition to its sparsity. In this method, the parsimony of model terms is evaluated using their locations in the prescribed candidate library, for the first time, considering the increased complexity with the power of polynomials and the order of spatial derivatives. Subsequently, the model parameters are updated through Bayesian inference with the raw data. This procedure aims to reduce the error associated with the possible loss of information in data preprocessing and numerical differentiation prior to sparse regression. Results of numerical case studies indicate that the governing PDEs of many canonical dynamical systems can be correctly identified using the proposed \textsc{Pe}SBL method from highly noisy data (up to 50\% in the current study). Next, the proposed methodology is extended for stochastic PDE learning where all parameters and modeling error are considered as random variables. Hierarchical Bayesian Inference (HBI) is integrated with the proposed framework for stochastic PDE learning from a population of observations. Finally, the proposed \textsc{PeSBL} is demonstrated for system response prediction with uncertainties and anomaly diagnosis. Codes of all demonstrated examples in this study are available on the website: \href{https://github.com/ymlasu}{\textcolor{blue}{https://github.com/ymlasu}}.

\end{abstract}

\end{frontmatter}
%\FloatBarrier
\section{Introduction}\label{Intro}
Despite that many dynamical systems can be well characterized by PDEs derived mathematically/physically from basic principles such as conservation laws, lots of other systems have unclear or elusive underlying mechanisms (e.g., ones in neuroscience, finance, and ecology). Thus, the governing equations are usually empirically formulated \cite{rudy2017data}. Data-driven physics discovery of dynamical systems gradually became possible in recent years due to the rapid development and extensive application of sensing technologies and computational power \cite{long2019pde}. Over the past years, extensive efforts have been devoted into discovering representative PDEs for complex dynamical systems of which limited prior knowledge are available \cite{schmidt2009distilling,raissi2018hidden,rudy2017data,long2019pde}. 

Among all the methods investigated for PDE identification/learning \cite{schmidt2009distilling,raissi2018hidden,rudy2017data,long2019pde,maslyaev1903data,atkinson2019data,hasan2020learning,xu2020dlga}, sparse regression gains the most attention in recent studies due to its inherent sparsity-promoting advantage. Considering a nonlinear PDE of the general form $u_t = N(u,u_x,u_{xx},...,x)$, in which the subscripts denote partial differentiation with respect to temporal or spatial coordinate(s), $N(\cdot)$ is an unknown expression on the right hand side of the PDE. $N(\cdot)$ is usually a nonlinear function of the spatial coordinate $x$, the measured quantity $u(x,t)$, and its spatial derivatives $u_x$ and $u_{xx}$. Given time series measurements of $u$ at certain spatial locations, the above equation can be approximated as $\mathbf{U}_t=\mathbf{\Theta}(\mathbf{U})\boldsymbol{\xi}$, in which $\mathbf{U}_t$ is the discretized form of $u_t$. $\mathbf{\Theta}(\mathbf{U})$ is a library matrix with each column corresponding to a candidate term in $N(\cdot)$. A key assumption in sparse identification is that $N(\cdot)$ consists of only a few term for a real physical system, which requires the solution of regression (i.e., $\xi$) to be a sparse vector with only a small number of nonzero elements. This assumption promotes a parsimonious form of the learned PDE and avoids overfitting the measured data with a complex model containing redundant nonlinear higher-order terms.

In most of current sparse regression methods for PDE learning, the parsimony of the learned model is realized through least squares regression with hard thresholding \cite{rudy2017data,rudy2019data,chen2020deep}, $\ell_1$ norm regularization \cite{bekar2021peridynamics,berg2019data,xiong2019data}, or $\ell_1$ norm regularization with hard thresholding \cite{both2021deepmod}. A critical challenge of these methods is that, the identification results using these method are susceptible to the selection of hyperparameters of the algorithm, including the regularizer $\lambda$ of the $\ell_1$ norm and the tolerance parameter $tol$ for hard thresholding. The identification results can be very different when hyperparameter settings change, especially when the measurement noise is large. As a result, the hyperparameter tuning is especially critical and challenging for cases with noisy measurements. Additionally, hard thresholding tends to suppress small coefficients that may not correspond to the most trivial terms of the intermediately learned PDEs. Recently, Zhang and Liu \cite{zhang2021robust} proposed a robust PDE learning method (i.e., the $\psi$-PDE method) that progressively selects important model terms from the candidate library which avoids the hyperparameter tuning. 

Above discussions are about the deterministic sparse regression. Probabilistic sparse regression has also been investigated for physics learning through Sparse Bayesian Learning (SBL) \cite{zhang2018robust,chen2021robust,fuentes2021equation,nayek2020spike,yuan2019machine,zhang2019robust,chen2020gaussian,Bhouri2021gaussian}. Unlike the deterministic sparse regression that yields a coefficient vector $\boldsymbol{\xi}$ without accompanying confidence intervals, the Bayesian approach enables quantifying uncertainties in the estimates of model parameters and enables model selection based on model complexities.  Another important benefit is that the evaluated model uncertainty in Bayesian inference can be propagated into the confidence levels in system predictions. It has been shown that the SBL method automatically avoids overfitting with a complex model through implementing a form of ``Occam's razor'' in the prescribed prior that controls the model complexity \cite{jefferys1992ockham}. Hence, the lemma of hyperparameter tuning can be largely obviated in SBL since it avoids setting additional regularization parameters \cite{tipping2003fast}.

Relevance Vector Machine (RVM) is the most widely used SBL method for Bayesian sparse regression in learning governing equations from observed data \cite{zhang2018robust,chen2021robust,fuentes2021equation,nayek2020spike,yuan2019machine,zhang2019robust}. RVM was originally developed for nonparametric modeling as a sparse and probabilistic alternative to the Support Vector Machine (SVM) \cite{tipping2003fast}. In RVM, sparsity is promoted by incorporating a parameterized independent Gaussian prior for each weight parameter, i.e., $p(w_i|\alpha_i) = \mathcal{N}(w_i|0,\alpha_i^{-1})$ with $w_i$ denoting a certain weight parameter and $\alpha_i$ its precision. The hyperparameter of each prior controls its strength over the associated weight. When solving this Bayesian inference problem through maximizing the evidence with respect to these hyperparameters, most of them go to infinity during the optimization. This leads to the posteriors of many weight parameters infinitely peaked at zero and consequently the weight vector $\mathbf{w}$ comprising only a few non-zero elements. RVM has been proved efficient for basis selection and advantageous over traditional methods (such as FOCUSS and basis pursuit) regarding obviating modeling/structural errors and algorithmic convergence \cite{wipf2004sparse}. These merits are essentially important in learning the governing equation(s) for a physical system, in which the model parsimony and interpretability must be preserved.

\begin{figure}[!h]
	\centering
	\includegraphics[scale=0.5]{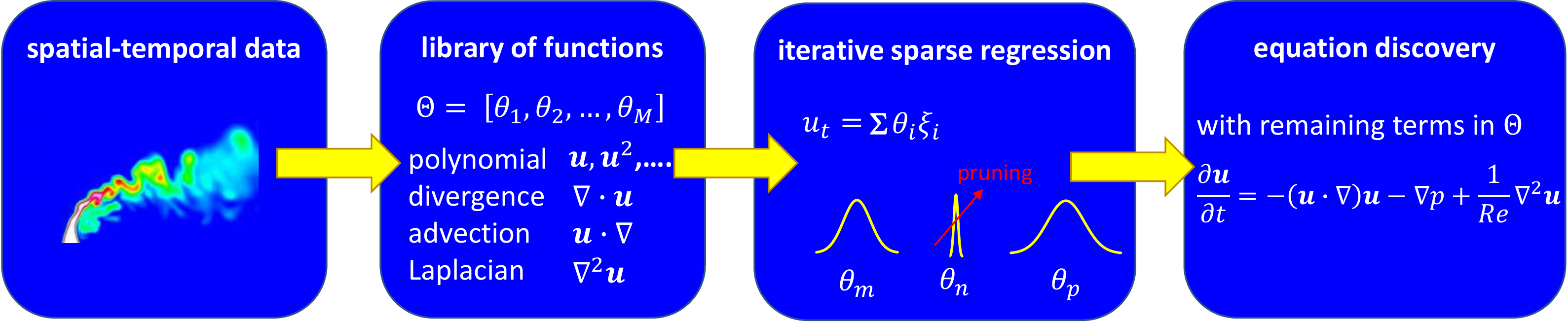}
	\caption{Framework of the SBL method for discovering PDE(s) from measured data.}
	\label{Figure:RVM}
\end{figure}

Figure \ref{Figure:RVM} illustrates the basic idea of discovering governing PDEs using the SBL method via RVM. With temporal/spatial derivatives numerically calculated from measured data in the spatial-temporal domain, a library for regression is built with representative terms containing physics meaning and/or others such as polynomials. Then the equation discovering problem can be defined as a sparse regression problem such as $\mathbf{U}_t=\mathbf{\Theta}(\mathbf{U})\boldsymbol{\xi}$ introduced above. This sparse regression problem can be solved sequentially (such as in \cite{tipping2003fast}), during which many terms in the prescribed library are excluded according to their statistics. Finally, the equations can be formulated using the remaining terms and their coefficients/weights. 

Fuentes et al. \cite{fuentes2021equation} used RVM to discover the governing Ordinary Differential Equations (ODEs) for several Single-Degree-of-Freedom (SDOF) nonlinear mechanical systems. Model uncertainty was quantified and propagated into predictions of system responses. This work was extended by Nayek et al. \cite{nayek2020spike} by adopting spike and slab priors for improved selective shrinkage of the weight parameters. Zhang and Lin \cite{zhang2018robust,zhang2019robust} attempted to discover ODEs/PDEs for the investigated dynamical systems using RVM. Robustness is lacking in this work because of the following limitations: 1) hard thresholding is implemented on the SBL results to further prune undisired terms, which compromises the merits of SBL methods for ODE/PDE learning; 2) as argued in \cite{lagergren2020learning}, noise is only added to the time derivative term instead of to the measured data, which largely suppresses the influence of noise in numerical differentiation and thus limits the method's robustness. This work was extended by Chen and Lin \cite{chen2021robust} through threshold Bayesian group Lasso with spike and slab prior (tBGL-SS) considering non-constant model parameters. However, the tBGL-SS method failed to resolve the hard thresholding issue. Additionally, the performance of PDE discovery significantly decreases when the noise level is beyond 5\%, which makes this method less robust than many state-of-the-art methods \cite{zhang2021robust} and thus limits its application in real practices.Yuan et al. \cite{yuan2019machine} used the RVM method to discover PDEs from measured spatiotemporal data and conducted convergence and consistency analysis. The robustness of this method in the presence of a large level of noise needs to be further examined. Zanna and Bolton \cite{zanna2020data} compared the SBL method using RVM and the physics-constrained deep learning (PCDL) using convolutional neural networks (CNN) for modeling turbulence in ocean mesoscale eddies and discussed the advantages and shortcomings of each method. In contrast to the PCDL method that yields a deep black-box CNN model, the SBL method provides physics-interpretable governing equations for the investigated complex dynamical system. 

Considering the limitations of existing studies on PDE learning via SBL approaches, this study proposes the Parsimony-Enhanced Sparse Bayesian Learning (\textsc{Pe}SBL) method for robust data-driven discovery of PDEs. The main argument in the proposed study is that the Parsimony principle should include both sparsity and complexity measures. Sparsity refers to the number of terms in a model, which has been widely addressed in the open literature. Complexity refers to the order and/or functional form complexity of each term, which has not been addressed in the physics discovery, to the best knowledge of the authors. Compared with traditional SBL methods for sparse regression, the proposed \textsc{Pe}SBL method enhances the parsimony of learned equations through quantifying model complexity and recommending less complex terms to the solution of sparse regression. The key issues investigated in this study include: 

\begin{enumerate}
	\item Preprocessing measured signals to improve the robustness of SBL in the presence of significant levels of noise;
	\item Promoting parsimony (rather than sparsity alone) of the identified physics model in PDE learning; 
	\item Quantifying uncertainties of the learned model and its representativeness of the intrinsic physics underlying observations; 
	\item Further reducing the uncertainties of learned model from SBL via Bayesian Model Updating (BMU);
	\item Propagating the learned model uncertainties into predictions of system responses; 
	\item Investigating system diagnosis and prognosis when system varies from time to time;
	\item Multiscale modeling of stochastic dynamical systems through Hierarchical Bayesian Inference (HBI); 
\end{enumerate}

The remaining part of this paper is structured as follows. Section \ref{Sec:method} establishes the framework of the \textsc{Pe}SBL method for discovering PDEs from observed dynamical systems; section \ref{Sec:results} presents and discusses the results of discovering governing equations for several canonical systems using the \textsc{Pe}SBL method; Section \ref{uncertainProp} investigates propagating the learned model uncertainties to the prediction of system dynamics; Section \ref{Section:DAP} investigates system diagnosis and prognosis when the essential properties of a certain system vary with time; Section \ref{Section:MSM} presents the framework and results of multiscale modeling and physics discovery through HBI;  Section \ref{Section:Conclusion} concludes this study with remarks and recommendations for future work.

%\FloatBarrier
\section{Methodology: the \textsc{Pe}SBL method for PDE learning}\label{Sec:method}
This section establishes the framework of the the \textsc{Pe}SBL method for learning governing PDEs from observed dynamical systems. The \textsc{Pe}SBL method is developed based on the RVM method of sparse Bayesian modeling or SBL. This method further promotes model parsimony instead of only sparsity when solving the regression problem sequentially through marginal likelihood maximization. Section \ref{Section:RVM} introduces the basic knowledge of SBL and RVM method; Section \ref{Section:PeSBL} presents the framework of \textsc{Pe}SBL, including the preprocessing procedures for minimizing the influence of measurement noise (Section \ref{sec:preprocess}), the sequential \textsc{Pe}SBL algorithm (Section \ref{Sec:PeSBL_Alg}) (with its robustness examined in Section \ref{Sec:robustness}), and Bayesian model updating (BMU) for further enhancing the performance of PDE learning using the measured raw data (Section \ref{Sec:BMU}).

\subsection{SBL and RVM \cite{tipping2003fast}}\label{Section:RVM} 

Given the measured dataset $\mathbf{t}=\left(t_{1}, \ldots, t_{N}\right)^{\mathrm{T}}$, it is assumed that $\mathbf{t}$ can be expressed as the sum of model approximation $\mathbf{y}=\left(y_1, \ldots, y_N\right)^{\mathrm{T}}$ and the residual $\epsilon=\left(\epsilon_{1}, \ldots, \epsilon_{N}\right)^{\mathrm{T}}$, such that
\begin{equation}
\mathbf{t} =\mathbf{y}+\boldsymbol{\epsilon} 
\end{equation}
In sparse regression, the model approximation $\mathbf{y}$ is expressed as the product of the library matrix $\mathbf{\Phi}$ and the coefficient vector $\mathbf{w}$, i.e.,
\begin{equation}
\mathbf{y} = \mathbf{\Phi} \mathbf{w}
\end{equation}
in which the library matrix $\mathbf{\Phi}=\left[\phi_{1} \ldots \phi_{M}\right]$ has a dimension of $N\times M$ with its columns comprising the over-complete set of $M$ basis vectors.

In sparse Bayesian modeling, the residuals $\epsilon_n(n=1,2,\ldots,N)$ are assumed following independent zero-mean Gaussian distributions, with a common variance $\sigma^2$, such that
\begin{equation}
\begin{split}
 p(\boldsymbol{\epsilon})&= \prod_{n=1}^{N}p(\epsilon_n)\\
 &=\prod_{n=1}^{N} \mathcal{N}\left(\epsilon_{n}|0,\sigma^{2}\right)
 \end{split}
 \end{equation}
Apparently, this assumption leads to a multivariate Gaussian likelihood for $\mathbf{t}$:
\begin{equation}\label{Eq:llh0}
p\left(\mathbf{t}|\mathbf{w}, \sigma^{2}\right)=(2 \pi)^{-N / 2} \sigma^{-N} \exp \left[-\frac{(\mathbf{t}-\mathbf{y})^{2}}{2 \sigma^{2}}\right]
\end{equation}
To promote sparsity, the coefficient parameters $\mathbf{w} = \left[w_1,\ldots,w_M\right]^\mathrm{T}$ are given independent Gaussian priors:
\begin{equation}\label{Eq:prior}
\begin{split}
p(\mathbf{w}|\boldsymbol{\alpha})&=\prod_{m=1}^{M}p(w_m|\alpha_m)\\
&=\prod_{m=1}^{M} \mathcal{N}\left(w_{m}|0,\alpha_{m}^{-1}\right)\\
&=(2 \pi)^{-M / 2} \prod_{m=1}^{M} \alpha_{m}^{1 / 2} \exp \left(-\frac{\alpha_{m} w_{m}^{2}}{2}\right)
\end{split}
\end{equation}
in which $\boldsymbol{\alpha}=\left(\alpha_{1},\ldots,\alpha_{M}\right)^{\mathrm{T}}$. Each element (e.g., $\alpha_m$) individually regularizes the strength of the prior over the corresponding coefficient parameter (e.g., $w_m$). 

Following the Bayes' rule, combining the prior in Equation \ref{Eq:prior} with the multivariate Gaussian likelihood in Equation \ref{Eq:llh0} yields the posterior distribution of $\mathbf{w}$: 
\begin{equation}
p\left(\mathbf{w}|\mathbf{t}, \boldsymbol{\alpha}, \sigma^{2}\right)=\frac{p\left(\mathbf{t}|\mathbf{w}, \sigma^{2}\right) p(\mathbf{w}|\boldsymbol{\alpha})}{p\left(\mathbf{t}|\boldsymbol{\alpha}, \sigma^{2}\right)}
\end{equation}
The posterior is proved to be Gaussian, that is,  $p\left(\mathbf{w}|\mathbf{t}, \boldsymbol{\alpha}, \sigma^{2}\right)=\mathcal{N}(\boldsymbol{\mu}, \mathbf{\Sigma})$ with
\begin{equation}\label{Eq:SM}
\mathbf{\Sigma}=\left(\mathbf{A}+\sigma^{-2} \mathbf{\Phi}^{\mathrm{T}} \mathbf{\Phi}\right)^{-1} \quad \boldsymbol{\mu}=\sigma^{-2} \mathbf{\Sigma} \mathbf{\Phi}^{\mathrm{T}} \mathbf{t}
\end{equation}
in which $\mathbf{A} = \mathrm{diag}(\alpha_1,...,\alpha_M)$. A most-probable point estimate of $\boldsymbol{\alpha}$, i.e., $\boldsymbol{\alpha}_\mathrm{MP}$, can be achieved by maximizing the following marginal log-likelihood $\mathcal{L}(\boldsymbol{\alpha})$ with respect to $\boldsymbol{\alpha}$:
\begin{equation} \label{Eq:llh}
\begin{split}
\mathcal{L}(\boldsymbol{\alpha}) &=\log p\left(\mathbf{t}|\boldsymbol{\alpha}, \sigma^{2}\right)=\log \int_{-\infty}^{\infty} p\left(\mathbf{t}|\mathbf{w}, \sigma^{2}\right) p(\mathbf{w}|\boldsymbol{\alpha}) d \mathbf{w} \\ &=-\frac{1}{2}\left[N \log 2 \pi+\log |\mathbf{C}|+\mathbf{t}^{\mathrm{T}} \mathbf{C}^{-1} \mathbf{t}\right] 
\end{split}
\end{equation}
in which
\begin{equation}
\mathbf{C}=\sigma^{2} \mathbf{I}+\mathbf{\Phi} \mathbf{A}^{-1} \mathbf{\Phi}^{\mathrm{T}}
\end{equation}

Subsequently, $\mathbf{\Sigma}$ and $\boldsymbol{\mu}$ can be estimated correspondingly from Equation \ref{Eq:SM} by setting $\boldsymbol{\alpha} = \boldsymbol{\alpha}_\mathrm{MP}$. When solving this optimization problem sequentially, it can be found that many elements of $\boldsymbol{\alpha}$ go to infinity, leading to the corresponding parameter posteriors peaked at zero infinitely and thus a sparse regression  model. However, sparsity itself may not be sufficient for learning governing equations from dynamical systems, which requires representing the intrinsic physics underlying measured data with the least number of simplest terms in the prescribed library without considerable loss of approximation accuracy. Therefore, the parsimony of the learned model must be guaranteed when defining or solving the regression problem through Bayesian inference. 

\subsection{The \textsc{PeSBL} Method} \label{Section:PeSBL}

\begin{figure}[!h]
	\centering
	\includegraphics[scale=0.5]{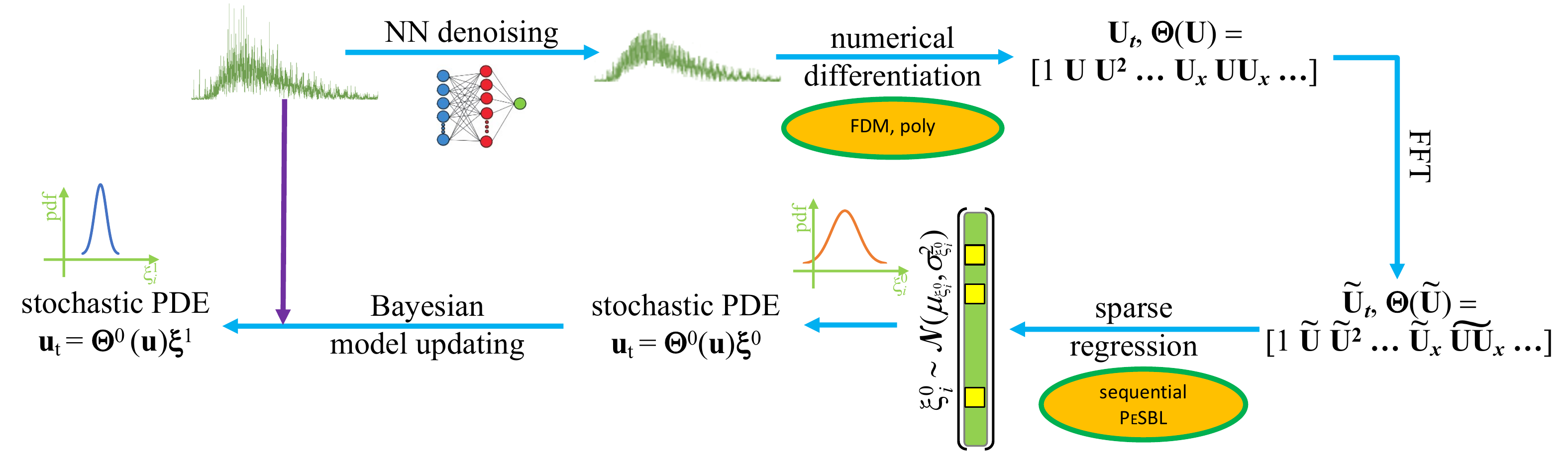}
	\caption{Framework of the \textsc{Pe}SBL method for discovering PDE(s) from measured data.}
	\label{Figure:Framework}
\end{figure}

Figure \ref{Figure:Framework} illustrates the framework of the proposed \textsc{Pe}SBL method for discovering the governing equations using measured data from a dynamical system. This framework starts from the noisy curve on the upper left corner, which denotes the measured signals containing all the information one can directly obtain from the instrumented system. The rest of this section will explain in detail each step of this framework.
\subsubsection{ Signal Preprocessing and Data Preparation} \label{sec:preprocess}
For preprocessing the measured data, a neural network (NN) model is built following the practices in \cite{xu2019dl,berg2019data}, setting the independent variables (i.e., $t$, $x$, etc.) as inputs and the measured quantity (e.g., $u$) as the output. The measured data are split into training and validation sets and the early stopping strategy is devised in the model training to prevent the NN model from overfitting the measured noise. A smoothed series of signals is expected from this preprocessing, which will be shown as follows taking the Burgers equation as an example.

\begin{figure}[!h]
	\centering
	\includegraphics[scale=0.8]{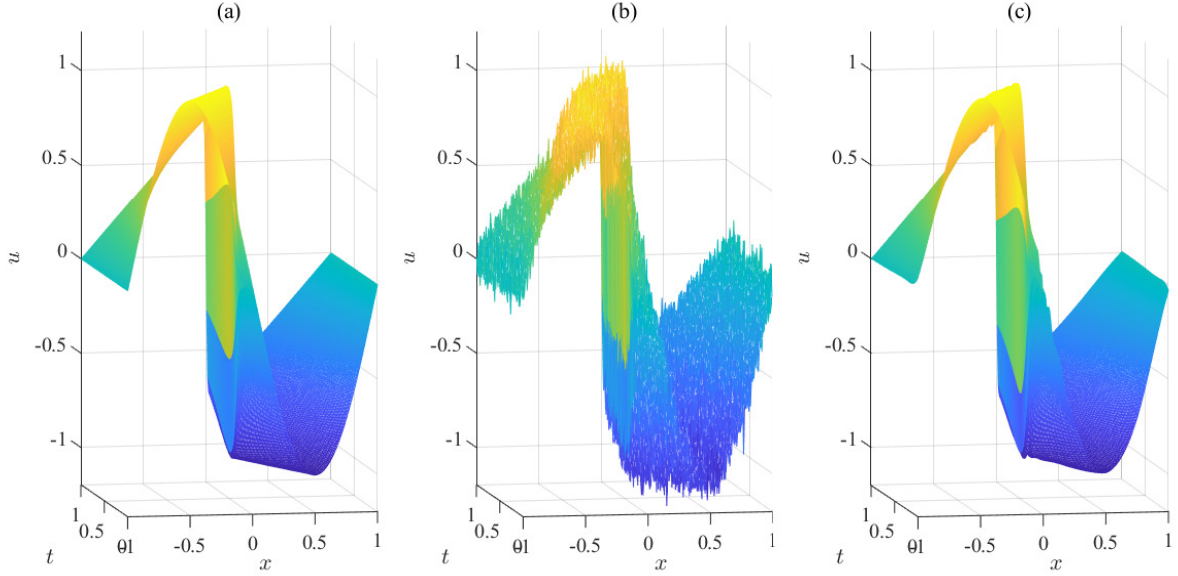}
	\caption{Simulated data for the 1D dissipative system characterized by the Burgers equation $u_t = -uu_x+\frac{0.01}{\pi} u_{xx}$ with (a) 0\% noise, (b) 10\% noise, and (c) 10\% noise afer NN denoising.}
	\label{Figure:solBurgers0}
\end{figure}

Burgers equation is used to describe the dynamics of a dissipative system. A 1D viscous Burgers equation is used to demonstrate the effects of signal preprocessing in the \textsc{PeSBL} method. It has the expression of $u_t = -uu_x+\nu u_{xx}$ with the initial conditions $u(0,x) = -\mathrm{sin}(\pi x)$ and boundary condition $u(t,-1) = u(t,1) =0$. In the Burgers equation, $\nu = \frac{0.01}{\pi}$ denotes the diffusion coefficient. It should be noted that compared with the Burgers equation widely used in the literature \cite{chen2021robust,rudy2017data} (i.e., $u_t = -uu_x+0.1u_{xx}$), this Burgers equation has a much smaller diffusion coefficient and thus is much more difficult to be correctly identified. In this study, this challenging equation is selected to examine the effectiveness and robustness of the proposed \textsc{PeSBL} method for PDE learning. Figure \ref{Figure:solBurgers0} (a) shows the simulated data from this dissipative system within the range $t\in [0,1]$ and $x\in [-1,1]$.

To demonstrate the effects of NN denoising step in the \textsc{PeSBL} method, 10\% white Gaussian noise is added to the numerical solution of the Burgers equation, which significantly varies the values of the solution (as shown in Figure \ref{Figure:solBurgers0} (b)) and thus poses challenge to calculating the numerical derivatives in the following step. In this study, the noise level is quantified by the percentage of the standard deviation of the measured variable. For example, if 10\% noise is added to $u$, then the outcome is $u_n = u+10\%*\mathrm{std}(u)*\mathrm{randn}(\mathrm{size}(u))$ where $\mathrm{std}(\cdot)$ evaluates the standard deviation, $\mathrm{randn}(\cdot)$ generates white Gaussian noise of the specified dimension, and $\mathrm{size}(\cdot)$ measures the dimension. Without much prior knowledge about the noise characteristics, NN modeling is applied to denoise the noisy measurements, and the processed data is visualized in Figure \ref{Figure:solBurgers0} (c). Comparing the three plots in Figure \ref{Figure:solBurgers0}, one can observe that NN denoising largely reduces the noise level in the collected data (from 10\% to 2\%) and makes the solution curve much smoother than the noisy measurement. Hence, it has the potential of improving the accuracy of subsequent numerical differentiation.

The denoised data will be subsequently used to calculate the numerical derivatives (i.e., $\mathbf{U}_t$, $\mathbf{U}_x$, $\mathbf{U}_{xx}$, etc.) and then construct the library matrix $\mathbf{\Theta}(\mathbf{U})$ for sparse regression. Numerical methods such as finite difference method (FDM) and polynomial interpolation are used to calculate the temporal and spatial derivatives. For the system characterized by the Burgers equation, the library is built with polynomials of $u$ to the power of 3, spatial derivatives to the $3^\mathrm{rd}$ order, and their products. As a result, it contains 16 terms in total, i.e., $\mathbf{\Theta}(\mathbf{U}) = \{1, \mathbf{U}, \mathbf{U}^2, \mathbf{U}^3, \mathbf{U}_x, \mathbf{UU}_x,...,  \mathbf{U}^3\mathbf{U}_{xxx}\}$. 

\begin{figure}[!h]
	\centering
	\includegraphics[scale=0.8]{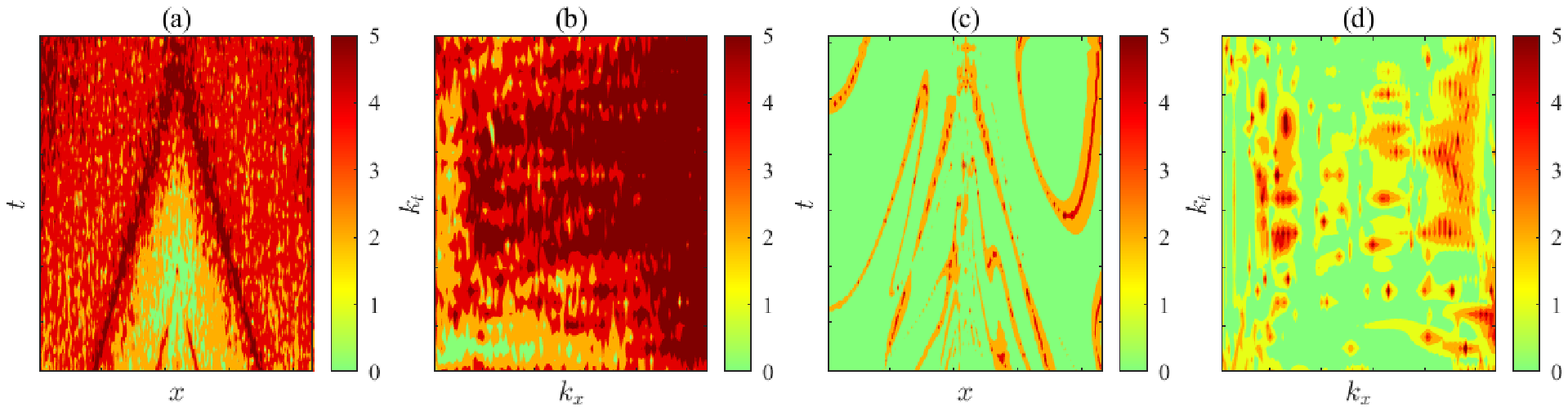}
	\caption{Color maps of relative errors in $u_{xx}$ of Burger equation caused by adding 50\% noise. 
		(a) $\mathrm{log}(\left|\frac{u^n_{xx}-u^0_{xx}}{u^0_{xx}}\right|)$;
		(b) $\mathrm{log}(\left|\frac{\widetilde{u}^n_{xx}-\widetilde{u}^0_{xx}}{\widetilde{u}^0_{xx}}\right|)$;
		(c) $\mathrm{log}(\left|\frac{u^{nn}_{xx}-u^0_{xx}}{u^0_{xx}}\right|)$;
		(d)$\mathrm{log}(\left|\frac{\widetilde{u}^{nn}_{xx}-\widetilde{u}^0_{xx}}{\widetilde{u}^0_{xx}}\right|)$. $u^0_{xx}$, $u^n_{xx}$, and $u^{nn}_{xx}$ are the $2^\mathrm{nd}$ order derivatives calculated using the clean data, noise data, and NN-denoised data, respectively; the tilde $\widetilde{(\cdot)}$ denote the results of 2D FFT; $k_x$ and $k_t$ represent the corresponding coordinates in the frequency domain.}
	\label{Figure:effectFFT}
\end{figure}

With $\mathbf{U}_t$ and $\mathbf{\Theta}(\mathbf{U})$ established, to further reduce the influence of noise in sparse regression, fast Fourier transform (FFT) is applied to transform $\mathbf{U}_t$ and $\mathbf{\Theta}(\mathbf{U})$ to their frequency domain counterparts $\widetilde{\mathbf{U}}_t$ and $\mathbf{\Theta}(\widetilde{\mathbf{U}})$, respectively. Following FFT, a frequency cutoff is implemented to preserve only the low frequency components that are expected to be less susceptible to noise. Moreover, this step converts the regression problem from the temporal-spatial domain to the frequency domain, which does not change the form of learned PDEs \cite{cao2020machine,zhang2021robust}. 

Taking the Burgers equation above as an example, Figures \ref{Figure:effectFFT} (a) to (d) compare the relative errors in the spatial-temporal domain and frequency domain. Figures \ref{Figure:effectFFT} (a) and (b) show the difference between the polluted data with 50\% noise and the simulated clean data. It can be observed that after taking 2D FFT, the low-frequency components are less affected by the added noise. In addition, Figures \ref{Figure:effectFFT} (c) and (d) show that NN denoising can largely reduce not only the relative error in the spatial-temporal domain, as can be predicted from Figure \ref{Figure:solBurgers0}, but also the relative error in the frequency domain. Therefore, it can be expected that the performance of PDE learning can be considerably improved by implementing these preprocessing procedures. This improvement will be demonstrated through comparison in Section \ref{Sec:results}.

\subsubsection{Sparse Regression Using the Sequential \textsc{PeSBL} Algorithm}\label{Sec:PeSBL_Alg}
With $\widetilde{\mathbf{U}}_t$ and $\mathbf{\Theta}(\widetilde{\mathbf{U}})$ from FFT with frequency cutoff, sparse regression is conducted using the sequential \textsc{PeSBL} algorithm proposed in this study. The details of this algorithm are elaborated in Algorithm \ref{alg1}. Unlike the sequential SBL algorithm in \cite{tipping2003fast}, the principle of this algorithm is promoting model parsimony instead of sparsity alone. This is achieved by preventing the algorithm from selecting/adding complex terms for marginal increase of the log-likelihood (i.e., $\mathcal{L}$ in Equation \ref{Eq:llh}). To this end, when establishing the library $\mathbf{\Theta}(\mathbf{U})$ for sparse regression, all terms in the library are arranged in an increasing level of complexity regarding the power of polynomials and the order of spatial derivatives. For example, for the Burgers equation, $\mathbf{\Theta}(\mathbf{U}) = \{1, \mathbf{U}, \mathbf{U}^2, \mathbf{U}^3, \mathbf{U}_x, \mathbf{UU}_x,...,  \mathbf{U}^3\mathbf{U}_{xxx}\}$. In this way, the index of a certain term or its function can be taken as an indicator of the complexity of this term, which is inspired by the concept of Minimum Description Length (MDL) that favors a short code for describing objects \cite{von2011statistical}. This study selects the square of the index of certain term in the library to evaluate its complexity, which is proved efficient for all the investigated dynamical systems. 

To incorporate model complexity in the evaluation of regression quality in each iteration of this sequential algorithm, the Akaike Information Criterion (AIC) for model selection \cite{wagenmakers2004aic} is modified as $\mathcal{\widetilde{AIC}}$ and defined as follows:
\begin{equation}
\mathcal{\widetilde{AIC}} = 2\frac{\Sigma i_s^{\scriptscriptstyle{2}}}{M}+2\mathrm{len}(i_s)-2\mathcal{L}
\end{equation}
in which $i_s$ denotes the list of indices of currently selected terms from the library, $\mathrm{len}(i_s)$ evaluates its length, and $M$ denotes the number of terms in the library, as in Section \ref{Section:RVM}. Compared with the standard AIC ($AIC = 2\mathrm{len}(i_s)-2\mathcal{L}$), $\mathcal{\widetilde{AIC}}$ penalizes the complexity of selected model terms in addition to the complexity from the number of terms in the current model. In addition, in the sequential \textsc{PeSBL} algorithm, the relative increase of log-likelihood is evaluated in each iteration. If it goes below a certain threshold (i.e., $tol_2$ in Algorithm \ref{alg1}), the operation of adding new terms will be obviated in this iteration (Line 34 in Algorithm \ref{alg1}). To the best of the authors' knowledge, this is the first time that parsimony is enhanced in regression problems to pursue the most representative model of the intrinsic physics underlying observed data. More details about this algorithm can be found in Algorithm \ref{alg1}.

Given the measured data from a certain dynamical system, with $\widetilde{\mathbf{U}}_t$ and $\mathbf{\Theta}(\widetilde{\mathbf{U}})$ taken as the inputs and the parameters (i.e., $maxIters$, $tol_1$, and $tol_2$) set, the sequential \textsc{PeSBL} algorithm yields the library $\mathbf{\Theta}^\mathrm{0}$ composed of remaining contributive terms and the corresponding coefficient vector $\boldsymbol{\xi}^0$  with each element following a Gaussian distribution, i.e., $\xi_i^0 \sim \mathcal{N}\left(\mu_{\xi_i^0},\sigma_{\xi_i^0}^2\right)$. These outputs can be used to formulate a stochastic PDE governing the observed system, i.e., $\boldsymbol{u}_\mathrm{t} = \boldsymbol{\Theta}^0(\boldsymbol{u})\boldsymbol{\xi}^0$. In Section \ref{Sec:results}, it will be demonstrated that the correct model forms can be successfully identified for several canonical dynamical systems using the sequential \textsc{PeSBL} algorithm. For example, with the clean data (0\% noise) measured from the dissipative system characterized by the Burgers equation ($u_t = -uu_x+\frac{0.01}{\pi} u_{xx}$), the sequential \textsc{PeSBL} algorithm yields $\boldsymbol{\Theta}^0 = \left\lbrace uu_x,u_{xx}\right\rbrace$ with $\xi_1^0 \sim \mathcal{N}(-1.02,1.06)$ and $\xi_2^0 \sim \mathcal{N}(-\frac{0.022}{\pi},5.19\times 10^{-5})$. Hence, the resulting PDE is $u_t = -1.02(\pm 1.03)uu_x+\frac{0.022}{\pi}(\pm 7.20\times10^{-3})u_{xx}$. In this study, the coefficients of PDE terms are presented in the form of $\mu_{\xi_i}(\pm \sigma_{\xi_i})$, in which $\mu_{\xi_i}$ denotes the mean of certain coefficient $\xi_i$ and $\sigma_{\xi_i}$ denotes its standard deviation. It shows that the learned model has limited coefficient accuracy and a significant level of uncertainty, while the model form is identical to the truth. This lack of model accuracy and considerable model uncertainty may result from the similarity of terms in the library (such as $uu_x$ and $u^2u_x$ for the Burgers equation) and will lead to even larger error and uncertainty in system predictions. Therefore, Section \ref{Sec:BMU} will investigate further improving the model accuracy and its confidence level through Bayesian inference.

\newcommand{\norm}[1]{\left\lVert#1\right\rVert}
\begin{algorithm}
\scriptsize
	\caption{\scriptsize \textsc{Pe}SBL algorithm: $[\mathbf{\Theta}^\mathrm{0},\boldsymbol{\mu}_{\xi^0},\boldsymbol{\alpha}_{\xi^0}] = \textsc{PeSBL}(\mathbf{\Theta},\widetilde{\mathbf{U}}_t,maxIters=1000,tol_1=10^{-4},tol_2=10^{-2}$)}\label{alg1}
	\begin{algorithmic}[1]
		
		\State \textbf{Input}: library matrix $\mathbf{\Theta}$ (with the size $N\times M$), discretized temporal derivative $\widetilde{\mathbf{U}}_t$ (with the size $N\times 1$), the maximum number of iterations $maxIters$, tolerance $tol_1$ of the $\Delta\mathcal{L}$ for stopping the iteration, and tolerance ratio $tol_2$ of the $\Delta\mathcal{L}$ for stopping adding new terms. $\Delta\mathcal{L}$ is the change of log-likelihood ($\mathcal{L}$).
		\State \textbf{output}:	parsimonious library $\mathbf{\Theta}^\mathrm{0}$ containing contributive terms, and the corresponding coefficient vector $\boldsymbol{\xi}^0$ with the mean $\boldsymbol{\mu}_{\xi^0}$ and precision $\boldsymbol{\alpha}_{\xi^0}$.
		\State Normalize $\mathbf{\Theta}$ and $\widetilde{\mathbf{U}}_t$:
		$\mathbf{\Theta} = \mathbf{\Theta}/\norm{\mathbf{\Theta}}_2$,
		$\widetilde{\mathbf{U}}_t = \widetilde{\mathbf{U}}_t/||\widetilde{\mathbf{U}}_t||_2$.
		\State Let $\mathbf{t} = \widetilde{\mathbf{U}}_t$ and $\mathbf{\Phi} = \mathbf{\Theta}$.
		\State Initialize $\mathcal{L}_{rec} = \emptyset$. \textcolor{gray}{\texttt{\#} $\mathcal{L}_{rec}$ denotes the record of log-likelihood in each iteration.} 
		\State Initialize $\sigma^2 = \mathrm{var}(\mathbf{t})$, $\alpha = \mathrm{Inf}(M,1)$, and $i_{s} = \emptyset$. \textcolor{gray}{\texttt{\#} var($\cdot$) denotes the variance of a given vector; Inf($\cdot$) denotes a matrix of all infinite values; $i_{s}$ denotes the collection of selected term indices during the following iterations.}
		\State Initialize $S = [S_1,S_2,...,S_M]$ \& $Q = [Q_1,Q_2,...,Q_M]$ with
		$S_m = \phi_m^\mathrm{T}\mathbf{C}^{-1}\phi_m$ \& $Q_m = \phi_m^\mathrm{T}\mathbf{C}^{-1}\mathbf{t}$.  \textcolor{gray}{\texttt{\#} Equation (22) in \cite{tipping2003fast} }
		
		\For{$i = 1,2,...,maxIters$}
			{
				\State $s = S$; $q = Q$;
				\State $s(i_{s}) = \frac{\alpha(i_{s}).*S(i_{s})}{\alpha(i_{s})-S(i_{s})}$; $q(i_{s}) = \frac{\alpha(i_{s}).*Q(i_{s})}{\alpha(i_{s})-S(i_{s})}$; \textcolor{gray}{\texttt{\#} Equation (23) in \cite{tipping2003fast}; $.*$ denotes elementwise multiplication. }
				\State $\theta = q^2-s$; \textcolor{gray}{\texttt{\#} Step 5 of the Sqential SBL Algorithm in Section 4 in \cite{tipping2003fast}}
				\State $i_{a} = (\theta>0)$; \textcolor{gray}{\texttt{\#} active indices in the current iteration, see Steps 6 \& 7 in \cite{tipping2003fast} }
				\State $i_{r} = i_{s}\cap i_{a}$; \textcolor{gray}{\texttt{\#} indices of terms to be re-estimated, see Steps 6 in \cite{tipping2003fast} }
				\State $i_{\scriptscriptstyle{+}} = i_{a}\setminus i_{s}$; \textcolor{gray}{\texttt{\#} indices of terms to be added, see Steps 7 in \cite{tipping2003fast} }
				\State $i_{\scriptscriptstyle{-}} = i_{s}\setminus i_{a}$; \textcolor{gray}{\texttt{\#} indices of terms to be deleted, see Steps 8 in \cite{tipping2003fast} }
				
				\State Initialize $\Delta\mathcal{L} = -\mathrm{Inf}(M,1)$; \textcolor{gray}{\texttt{\#} potential change of log-likelihood}
				\State Initialize $\Delta\mathcal{C} = \mathrm{O}(M,1)$; \textcolor{gray}{\texttt{\#} potential change of model complexity $\mathcal{C}$;  $\mathrm{O}$ denotes a matrix of all zero values.}
				\State Initialize $\mathcal{C}_1 = \mathrm{O}(M,1)$; \textcolor{gray}{\texttt{\#} potential changed model complexity}
				\State $\mathcal{C}_0 = 2\Sigma i_s^{\scriptscriptstyle{2}}/M+2\mathrm{len}(i_s)$; \textcolor{gray}{\texttt{\#} initial model complexity; $\mathrm{len}(\cdot)$ denotes the length of a vector.}
				\If{$i_{r}\neq\emptyset$}
				\State $\widetilde{\alpha} = \frac{s(i_{r})}{\theta(i_{r})}$; \textcolor{gray}{\texttt{\#} Equation (20) in  \cite{tipping2003fast} }	
				\State $d\alpha = \widetilde{\alpha}^{-1}-\alpha(i_{r})^{-1}$;
				\State$2\Delta\mathcal{L}(i_{r}) = \frac{Q(i_{r})^2}{S(i_{r})+ {d\alpha}^{-1}}-\mathrm{log}\left[1 + S(i_{r})d\alpha \right]$;
				 \textcolor{gray}{\texttt{\#} Equation (32) in  \cite{tipping2003fast} }			
				\EndIf
				
				\If{$i_{\scriptscriptstyle{+}}\neq\emptyset$}
				\State$2\Delta\mathcal{L}(i_{\scriptscriptstyle{+}}) = \frac{Q(i_{\scriptscriptstyle{+}})^2-S(i_{\scriptscriptstyle{+}})}{S(i_{\scriptscriptstyle{+}})}+\mathrm{log}\frac{S(i_{\scriptscriptstyle{+}})}{Q(i_{\scriptscriptstyle{+}})^2}$;  \textcolor{gray}{\texttt{\#} Equation (27) in  \cite{tipping2003fast} }	
				\State $\mathcal{C}_1(i_{\scriptscriptstyle{+}}) = 2(\Sigma i_s^{\scriptscriptstyle{2}}+i_{\scriptscriptstyle{+}}^{\scriptscriptstyle{2}})/M+2(len(i_s)+1)$; 
				\State $\Delta\mathcal{C}(i_{\scriptscriptstyle{+}}) = \mathcal{C}_1(i_{\scriptscriptstyle{+}})-\mathcal{C}_0$;
				\EndIf
				
				\If{$i_{\scriptscriptstyle{-}}\neq\emptyset$}
				\State$2\Delta\mathcal{L}(i_{\scriptscriptstyle{-}}) = \frac{Q(i_{\scriptscriptstyle{-}})^2}{S(i_{\scriptscriptstyle{-}})-\alpha(i_{\scriptscriptstyle{-}})}-\mathrm{log}\left(1-\frac{S(i_{\scriptscriptstyle{-}})}{\alpha(i_{\scriptscriptstyle{-}})}\right)$;  \textcolor{gray}{\texttt{\#} Equation (37) in  \cite{tipping2003fast} }	
				\State $\mathcal{C}_1(i_{\scriptscriptstyle{-}}) = 2(\Sigma i_s^{\scriptscriptstyle{2}}-i_{\scriptscriptstyle{-}}^{\scriptscriptstyle{2}})/M+2(\mathrm{len}(i_s)-1)$; 
				\State $\Delta\mathcal{C}(i_{\scriptscriptstyle{-}}) = \mathcal{C}_1(i_{\scriptscriptstyle{-}})-\mathcal{C}_0$;
				\EndIf
				\State $\Delta \mathcal{\widetilde{AIC}} = \Delta\mathcal{C}-2\Delta\mathcal{L}$; \textcolor{gray}{\texttt{\#}  potential change of $\mathcal{\widetilde{AIC}}$}
				\State $\left[\Delta\mathcal{\widetilde{AIC}}_{m},i_{\scriptscriptstyle{m}}\right] = min\left(\Delta\mathcal{\widetilde{AIC}}\right)$; \textcolor{gray}{\texttt{\#} find the operation yielding the smallest $\Delta\mathcal{\widetilde{AIC}}$}
				
				\textcolor{gray}{\texttt{\#}  If the relative increase of log-likelihood is smaller than $tol_2$, then obviate adding new terms in this iteration.}
				\If{$\Delta\mathcal{L}(i_{\scriptscriptstyle{m}})<tol_2*\mathcal{L}(i_1) \;\&\; i_{\scriptscriptstyle{m}}\in i_{\scriptscriptstyle{+}}$}
				$\left[\Delta\mathcal{\widetilde{AIC}}_{m},i_{\scriptscriptstyle{m}}\right] = min\left(\Delta\mathcal{\widetilde{AIC}}(i_{\scriptscriptstyle{s}})\right)$;
				\EndIf
				\If{$\Delta\mathcal{L}(i_{\scriptscriptstyle{m}})<tol_1$} \textbf{break};
				\EndIf
				\State $\Delta\mathcal{L}_{m} = \Delta\mathcal{L}(i_{\scriptscriptstyle{m}})$; \textcolor{gray}{\texttt{\#} change of log-likelihood corresponding to $\Delta\mathcal{\widetilde{AIC}}_{m}$}
				\State $\mathcal{L}_{rec}(i) = \mathcal{L}_{rec}(i-1)+\Delta\mathcal{L}_{m}$;
				
				\textcolor{gray}{\texttt{\#} Update all variables according to the operation type.}
				\Switch{$i_{m}$}
   				\Case{$\in i_{r}$}
      				\State Update $\mathbf{\Sigma}$,$\boldsymbol{\mu}$,$S$,and $Q$ by Equations (33) to (36) in  \cite{tipping2003fast};
				\State Update $\alpha$ by  $\alpha(i_{m}) = \frac{s(i_{m})}{\theta(i_{m})}$; \textcolor{gray}{\texttt{\#} Equation (20) in  \cite{tipping2003fast} }	
   				\EndCase
    				\Case{$\in i_{\scriptscriptstyle{+}}$}
      				\State Update $\mathbf{\Sigma}$,$\boldsymbol{\mu}$,$S$,and $Q$  by Equations (28) to (31) in  \cite{tipping2003fast};
				\State Update $\alpha$ by  $\alpha(i_{m}) = \frac{s(i_{m})}{\theta(i_{m})}$; \textcolor{gray}{\texttt{\#} Equation (20) in  \cite{tipping2003fast} }	
				\State Update $i_s$ by $i_s = i_s \cup i_m$;
    				\EndCase
				\Case{$\in i_{\scriptscriptstyle{-}}$}
      				\State Update $\mathbf{\Sigma}$,$\boldsymbol{\mu}$,$S$,and $Q$ by Equations (38) to (41) in  \cite{tipping2003fast};    		
				\State Update $\alpha$ by  $\alpha(i_{m}) = \mathrm{Inf}$; 	
				\State Update $i_s$ by $i_s = i_s \setminus i_m$;	
				\EndCase
 				\EndSwitch
			}	
		\EndFor	
		 \State Outputs: $\mathbf{\Theta}^\mathrm{0} = \mathbf{\Theta}(:,i_s)$; $\boldsymbol{\mu}_{\xi^0} = \boldsymbol{\mu}$; $\boldsymbol{\alpha}_{\xi^0} = \alpha(i_s)$.

	\end{algorithmic}
\end{algorithm}

\subsubsection{Robustness of Algorithm \ref{alg1} Regarding Variations of Parameters $tol_1$ and $tol_2$}\label{Sec:robustness}
Many existing PDE learning methods in the literature, whether deterministic \cite{rudy2017data,both2021deepmod,chen2020deep} or probabilistic \cite{zhang2018robust,chen2021robust}, lack robustness in the learned model forms, because they intermediately or finally determine model terms through hard thresholding and thus easily fall into the lemma of hyperparameter tuning. Hence, it is of essential importance to examine the robustness of the proposed \textsc{PeSBL} method with respect to the variation of its hyperparameters (i.e., $tol_1$ \& $tol_2$ in Algorithm \ref{alg1}) in addition to the robustness with respect to the measurement noise. Without loss of generality, the Burgers equation ($u_t = -uu_x+\frac{0.01}{\pi}u_{xx}$) with 20\% measurement noise is used as an example in this analysis. Considering the challenge of learning this Burgers equation with such a small diffusion coefficient, the conclusion about the method robustness should be convincing and generalizes well to other system equations.

In Algorithm \ref{alg1},  $tol_1$ is the convergence criterion of the iteration regarding the increase of log-likelihood $\Delta\mathcal{L}$. When $\Delta\mathcal{L}$ goes below $tol_1$, the algorithm will stop updating the model form by re-estimating existing terms, adding new terms, or deleting unimportant terms and prepare the outputs. In this analysis, the value of $tol_1$ is varied from $10^{-6}$ to $10^{-1}$ to test its influence on the algorithm outcome. It is found that the correct terms of the Burgers equation (i.e., $uu_x$ and $u_{xx}$) can always be extracted without any redundant terms added. When a large value is assigned for $tol_1$, for example $10^{-1}$, the ``re-estimate'' iterations in Algorithm \ref{alg1} will be reduced, which will slightly affect the coefficient means and variances. Unlike $tol_1$, $tol_2$ prevents the algorithm from adding terms that will not considerably increase the log-likelihood but increases the model complexity. In this analysis, $tol_2$ is varied in the same range with $tol_1$. It shows that the condition in Line 34 of Algorithm \ref{alg1} is never triggered for the Burger equation, so that the variation of $tol_2$ does not affect the results of PDE learning. This means that the modeling complexity defined in Algorithm \ref{alg1} is sufficient to guarantee the proper parsimony of learned models. The effects of $tol_2$ will be further investigated for other systems in future studies. Therefore, compared with the current methods in the literature, the proposed \textsc{PeSBL} method has advantageous robustness in learning the correct model forms. This merit is noteworthy especially when faced with a novel system of which very limited prior knowledge is available.

\subsubsection{Bayesian Model Updating (BMU) with the Raw Data} \label{Sec:BMU}
Taking the Gaussian distributions of model parameters obtained from sparse regression as priors, their accuracy and confidence level can be further improved through BMU when more data/information are available. In this section, the measured raw data is reused in BMU, considering possible loss of information during signal processing and numerical differentiation. 

For the dynamical systems investigated in this study, the evidence for BMU is set as the system measurement $\widetilde{u}$. With the model form correctly identified from sparse regression, the system response (i.e., $u(\boldsymbol{\xi})$) with certain model parameter set (i.e., $\boldsymbol{\xi}$) can be estimated by solving the associated PDE(s) $\boldsymbol{u}_\mathrm{t} = \boldsymbol{\Theta}^0(\boldsymbol{u})\boldsymbol{\xi}$ numerically with  boundary/initial conditions given or extracted from measurements. Then the error function (i.e., the estimation error) can be defined as the discrepancy between the estimation $u(\boldsymbol{\xi})$) and measurement $\widetilde{u}$ normalized by the $\ell_2$ norm of $\widetilde{u}$, such that
\begin{equation}\label{Eq:err}
\boldsymbol{e} = \frac{\widetilde{u}-u(\boldsymbol{\xi})}{  \lVert \widetilde{u} \rVert}
\end{equation}

For the convenience of usage in BMU, $\boldsymbol{e}$ is reshaped to a column vector. This study assumes that the elements of $\boldsymbol{e}$ (i.e., $e_i$, $i=1,2,..., N_e$; $N_e$ is the number of error terms) are independent and follows an identical Gaussian distribution for the maximum information entropy. That is
\begin{equation}\label{Eq:err2}
\mathrm{i.i.d.} \quad e_i \sim \mathcal{N}(\mu_e,\sigma_e^2)
\end{equation}
which provides the likelihood function of the measurement $\widetilde{u}$. Given certain value of $\boldsymbol{\xi}$, the difference between the measurement $\widetilde{u}$ and prediction $u(\boldsymbol{\xi})$ follows a multivariate Gaussian distribution. That is,
\begin{equation}\label{Eq:llh}
\begin{split}
p(\widetilde{u}\big\lvert \boldsymbol{\xi},\mu_e,\sigma_e^2) &= p(\boldsymbol{e}\big\lvert \mu_e,\sigma_e^2) \\
&= \prod_{i=1}^{N_e}\mathcal{N}(e_i\big\lvert \mu_e,\sigma_e^2)
\end{split}
\end{equation}

In BMU, the priors of $\xi_i$ are set as their posteriors obtained from sparse regression in Section \ref{Sec:PeSBL_Alg}. The mean of errors $\mu_e$ is assumed following a zero-mean Gaussian distribution with its standard deviation equal to 1/3, such that most of error values lie in the range $(-1,1)$. An non-informative inverse-gamma prior is used for $\sigma_e^2$. In summary,
\begin{equation}\label{Eq:prior1}
\xi_i \sim \mathcal{N}\left(\mu_{\xi_i^0},\sigma_{\xi_i^0}^2\right)
\end{equation}
\begin{equation}\label{Eq:prior2}
\mu_e \sim \mathcal{N}(0,\sigma_{\mu_e}^2)
\end{equation}
\begin{equation}\label{Eq:prior3}
\sigma_e^2 \sim Inv\-Gamma(\alpha_e,\beta_e)
\end{equation}
in which $\sigma_{\mu_e}^2 = \left(\frac{1}{3}\right)^2 = \frac{1}{9}$, $\alpha_e=1$ is the shape parameters of the inverse-gamma distribution, and $\beta_e=2$ is its scale parameter. With the likelihood function and priors specified, according to the Bayes' theorem, posterior probability density function (PDF) of the parameter set (including $\boldsymbol{\xi} = \{\xi_1,\xi_2,...,\xi_M\}$, $\mu_e$, and $\sigma_e$) is proportional to the product of the likelihood function and the prior PDFs. That is:
 
\begin{equation}
p\left(\boldsymbol{\xi},\mu_e,\sigma_e^2 \big\lvert \widetilde{u}\right)
\propto 
p(\widetilde{u}\big\lvert \boldsymbol{\xi},\mu_e,\sigma_e^2) 
p(\boldsymbol{\xi})p(\mu_e)p(\sigma_e^2)
\end{equation}

Plugging in the expressions of priors in Equations \ref{Eq:prior1} to \ref{Eq:prior3} and likelihood function in Equation \ref{Eq:llh}, the posterior PDF becomes

\begin{equation}\label{Eq:postPDF1}
\begin{split}
&p\left(\boldsymbol{\xi},\mu_e,\sigma_e^2 \big\lvert \widetilde{u}\right)
\propto\\
&\left\lbrace\prod_{i=1}^{N_e}\frac{1}{\sqrt{\sigma_e^2}}\mathrm{exp}\left[-\frac{1}{2}\left(\frac{e_i-\mu_e}{\sigma_e}\right)^2\right]\right\rbrace
\left\lbrace\prod_{i=1}^{M}\frac{1}{\sqrt{\sigma_{\xi_i^0}^2}}\mathrm{exp}\left[-\frac{1}{2}\left(\frac{\xi_i-\mu_{\xi_i^0}}{\sigma_{\xi_i^0}}\right)^2\right]\right\rbrace\\
&\mathrm{exp}\left[-\frac{1}{2}\left(\frac{\mu_e}{\sigma_{\mu_e}}\right)^2\right]
(\sigma_e^2)^{-\alpha_e-1}\mathrm{exp}\left(-\frac{\beta_e}{\sigma_e^2}\right)\\
&\propto (\sigma_e^2)^{-\frac{N_e}{2}-\alpha_e-1}
\mathrm{exp}\left[-\frac{1}{2}\sum_{i=1}^{N_e}\left(\frac{e_i-\mu_e}{\sigma_e}\right)^2
-\frac{1}{2}\sum_{i=1}^M\left(\frac{\xi_i-\mu_{\xi_i^0}}{\sigma_{\xi_i^0}}\right)^2
-\frac{1}{2}\left(\frac{\mu_e}{\sigma_{\mu_e}}\right)^2
-\frac{\beta_e}{\sigma_e^2}\right]
\end{split}
\end{equation}

The posterior PDF in Equation \ref{Eq:postPDF1} is known hard to be solved analytically \cite{rubinstein2016simulation}. Among the numerical methods for solving the  posterior PDFs (e.g., variational inference, Markov Chain Monte Carlo (MCMC), and expectation-maximization (EM)), Gibbs sampling, a branch of MCMC algorithm, has been proved efficient in approximating multivariate probability distributions, especially when direct sampling from the PDF is challenging. Gibbs sampling requires deriving the full conditional distribution of each parameter from the joint posterior PDF. For certain parameters, their conditional posterior distributions are found to be in standard forms, such as normal distribution. In this case, these parameters can be directly sampled in each iteration. Otherwise, the Metropolis-Hasting algorithm can be used for sampling by proposing candidate parameter values and accepting or rejecting them according to certain criteria. If the Metropolis-Hasting algorithm is used for sampling certain parameter(s), the algorithm is usually called Metropolis-within-Gibbs \cite{rubinstein2016simulation}.  

The posterior full conditional distribution of a parameter (or a parameter set) is the conditional distribution of that parameter given current values of all other parameters. For a certain parameter $v$, the rest of parameters can be denoted as $V_{-v}$, then the full conditional distribution $P(v|V_{-v})$ has the form \cite{gilks1995markov}:
\begin{equation}\label{Eq:conditional}
\begin{split}
P(v|V_{-v})& \propto P(v,V_{-v})\\
  %& \propto \text{terms in } P(V) \text{ containing } v\\
  & = P(v|\text{parents}[v]) \prod_{w \in \text{children}[v]} P(w|\text{parents}[w])
\end{split}
\end{equation}
in which parents of a parameter are the parameters contributive to the generation of $v$ and children of a parameter denote parameters resulting from it. Following this principle, the full conditional distributions can be derived as follows:

\begin{equation} \label{eq:cod_pos1}
p\left(\xi_i\big\vert\cdot\right)
\propto
\mathrm{exp}\left[-\frac{1}{2}\sum_{i=1}^{N_e}\left(\frac{e_i-\mu_e}{\sigma_e}\right)^2
-\frac{1}{2}\left(\frac{\xi_i-\mu_{\xi_i^0}}{\sigma_{\xi_i^0}}\right)^2\right]
\end{equation}
\begin{equation} \label{eq:cod_pos2}
\begin{split}
p\left(\mu_e\big\vert\cdot\right)
&\propto
\mathrm{exp}\left[-\frac{1}{2}\sum_{i=1}^{N_e}\left(\frac{e_i-\mu_e}{\sigma_e}\right)^2
-\frac{1}{2}\left(\frac{\mu_e}{\sigma_{\mu_e}}\right)^2\right]\\
&\sim\mathcal{N}\left(\frac{\sum_{i=1}^{N_e}e_i}{N_e+\frac{\sigma_e^2}{\sigma_{\mu_e}^2}},\frac{1}{\frac{N_e}{\sigma_e^2}+\frac{1}{\sigma_{\mu_e}^2}}\right)
\end{split}
\end{equation}
\begin{equation}
\begin{split}
\left(\sigma_e^2\big\vert\cdot\right) 
&\propto
 (\sigma_e^2)^{-\frac{N_e}{2}-\alpha_e-1}
\mathrm{exp}\left[-\frac{1}{2}\sum_{i=1}^{N_e}\left(\frac{e_i-\mu_e}{\sigma_e}\right)^2
-\frac{\beta_e}{\sigma_e^2}\right]\\
&\sim Inv\-Gamma\left(\frac{N_e}{2}+\alpha_e,\frac{1}{2}\sum_{i=1}^{N_e}(e_i-\mu_e)^2+\beta_e\right)\\
\end{split}
\end{equation}

A Metropolis-within-Gibbs scheme is necessary for sampling the posterior PDFs of the parameter set, since the conditional posterior distribution of $\xi_i$ is analytically intractable while that of $\mu_e$ and $\sigma_e^2$ are both in standard form. The convergence of Markov chains can be determined as follows \cite{cowles1996markov}: 1) run several simulations independently with different random starting points; 2) check the ratio of inter-chain to intra-chain variances of all parameters; 3) if the ratio becomes very close to 1, then stationary distribution is achieved. 

\begin{figure}[!h]
	\centering
	\includegraphics[scale=0.8]{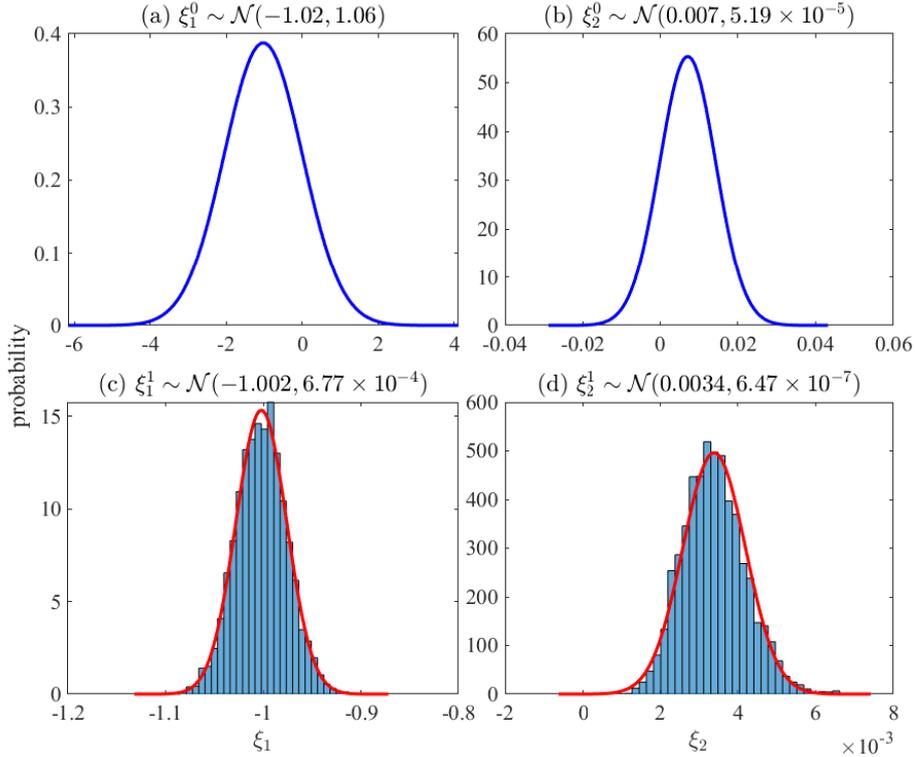}
	\caption{Results of sparse regression and BMU for the Burgers equation ($u_t = -uu_x+\frac{0.01}{\pi}u_{xx}$) with 0\% noise. $\xi_1$ and $\xi_2$ denote the coefficient of $uu_x$ and $u_{xx}$, respectively. The superscripts $^0$ and $^1$ denote the prior obtained from sparse regression using the sequential \textsc{PeSBL} algorithm and the posterior after BMU. (c) and (d) show the distribution of samples by Metroplis-within-Gibbs with histograms.}
	\label{Figure:BMU_burgersN0}
\end{figure}

In summary, Bayesian inference is conducted in this step to further update the model parameters obtained from sparse regression by exploiting the information contained in the raw data. It is expected to improve the accuracy and confidence level of the learned sparse model. Taking the Burgers equation ($u_t = -uu_x+\frac{0.01}{\pi}u_{xx}$) as an example, with the data containing 0\% noise, the PDE after BMU becomes $u_t = -1.0020(\pm 0.0260)uu_x+\frac{0.0107}{\pi}(\pm 8.04\times10^{-4})u_{xx}$. Comparing this model with that from sparse regression in Section \ref{Sec:PeSBL_Alg} (i.e., $u_t = -1.02(\pm 1.03)uu_x+\frac{0.022}{\pi}(\pm 7.20\times10^{-3})u_{xx}$), one can find that BMU considerably improves the accuracy of model coefficients and reduces the model's uncertainty. This comparison is more clearly demonstrated in Figure \ref{Figure:BMU_burgersN0}. The performance of the proposed \textsc{PeSBL} method in learning the correct governing PDEs will be further examined with more canonical dynamical systems in Section \ref{Sec:results}.

%\FloatBarrier
\section{Results of PDE Learning and Discussions}\label{Sec:results}
This section presents and discusses the results of PDE learning using the \textsc{PeSBL} method with simulated noisy data from several canonical systems covering a number of scientific domains. 0 to 50\% noise is added to the numerically simulated clean data to demonstrate the effects of preprocessing and the robustness of the \textsc{PeSBL} method in cases with noisy measurements. Two 1D systems are investigated in section \ref{results1D}: (1) the dissipative system characterized by the 1D Burgers equation (as shown in section \ref{Sec:method}); (2) the traveling waves described by the Korteweg–de Vries (KdV) equation. Section \ref{results2D} presents the results of two 2D systems: (1) an extended dissipative system characterized by the 2D Burgers equation; (2) the lid-driven cavity flow governed by the 2D Navier Stokes equation. Codes of all demonstrated examples are available on the website: \href{https://github.com/ymlasu}{\textcolor{blue}{https://github.com/ymlasu}}.
%\FloatBarrier
\subsection{Discovering PDEs for 1D systems}\label{results1D}
\begin{figure}[!h]
	\centering
	\includegraphics[scale=0.8]{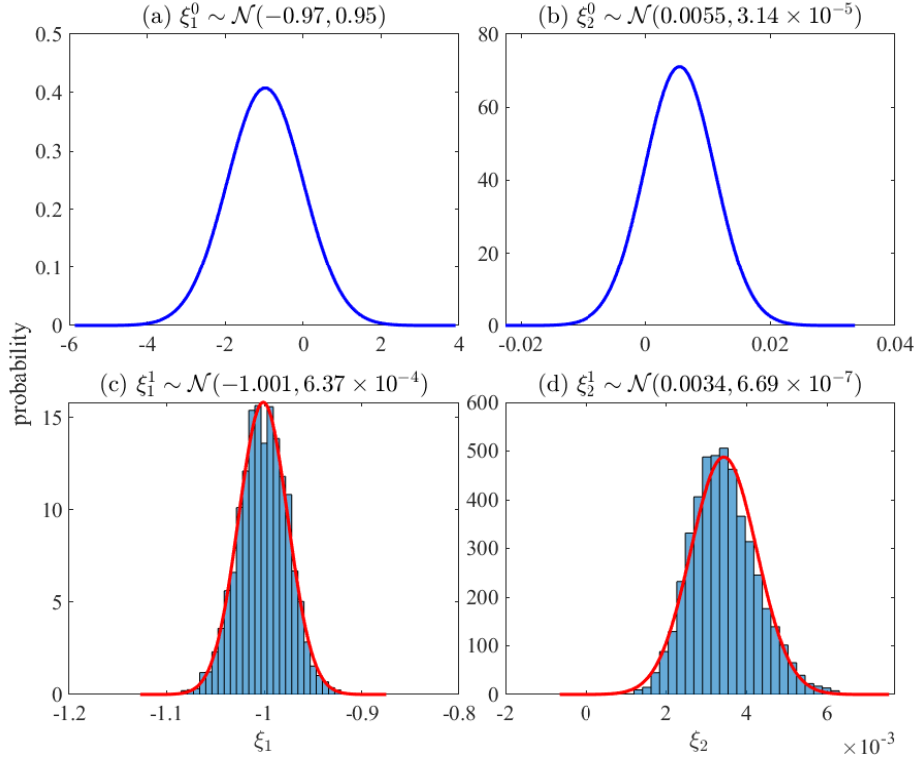}
	\caption{Results of sparse regression and BMU for the Burgers equation ($u_t = -uu_x+\frac{0.01}{\pi}u_{xx}$) with 20\% noise. Please refer to the caption of Figure \ref{Figure:BMU_burgersN0} for more information.}
	\label{Figure:BMU_burgersN20}
\end{figure}

Table \ref{Table:burgersNoisy} lists the results of learning the Burgers equation ($u_t = -uu_x+\frac{0.01}{\pi}u_{xx}$) from noisy data using the proposed \textsc{PeSBL} method. It shows that the \textsc{PeSBL} method yields identically correct PDE form with data containing up to 30\% noise. Figures \ref{Figure:BMU_burgersN20} (a) to (d) show the intermediate and final results of stochastic PDE learning using the \textsc{PeSBL} method with measured data containing 20\% noise from this dissipative system. It can be observed that, by virtue of the BMU step in the \textsc{PeSBL} method, the mean values of coefficients get very close to that of the ground truth PDE and the model uncertainty is largely reduced compared with the results of sparse regression. When the noise level goes above 30\%, the diffusion term of the Burgers equation (i.e., $\frac{0.01}{\pi}u_{xx}$) cannot be identified using the \textsc{PeSBL} method due to the challenge from the extremely small diffusion coefficient. It is worth noting that, the Burgers equation with a much larger diffusion coefficient (i.e., $u_t = -uu_x+0.1u_{xx}$) is widely used in existing studies \cite{chen2021robust,rudy2017data} for PDE learning. It has been verified that with the \textsc{PeSBL} method, the correct form of this less challenging Burgers equation can be successfully identified even with data containing larger than 50\% noise.
\begin{table} [!h]
	\caption{Results of PDE learning using the \textsc{PeSBL} method (Burgers equation: $u_t = -uu_x+\frac{0.01}{\pi}u_{xx}$).}
	\begin{tabular}{ |c|c| } 
		\hline
		noise level &  identified PDE \\ \hline
		0\% & $u_t = -1.0020(\pm 0.0260)uu_x+\frac{0.0107}{\pi}(\pm 8.04\times10^{-4})u_{xx}$\\ \hline
		10\% &  $u_t = -1.0007(\pm0.0255)uu_x+\frac{0.0106}{\pi}(\pm 7.98\times10^{-4})u_{xx}$\\ \hline		
		20\% & $u_t = -1.0010(\pm0.0252)uu_x+\frac{0.0108}{\pi}(\pm 8.18\times10^{-4})u_{xx}$\\ \hline
		30\% & $u_t = -0.9988(\pm0.0262)uu_x+\frac{0.0107}{\pi}(\pm 8.21\times10^{-4})u_{xx}$\\ \hline
%		40\% & $u_t = -0.9987(\pm 0.0260)uu_x+\frac{0.0108}{\pi}(\pm 8.52\times10^{-4})u_{xx}$\\ \hline
%		50\% &  $u_t = 0.0331(\pm 0.0487)u-1.0149(\pm 0.0377)uu_x+\frac{0.0123}{\pi}(\pm 0.0011)u_{xx}$\\ \hline
	\end{tabular}\label{Table:burgersNoisy}
\end{table}

\begin{figure}[!h]
	\centering
	\includegraphics[scale=0.8]{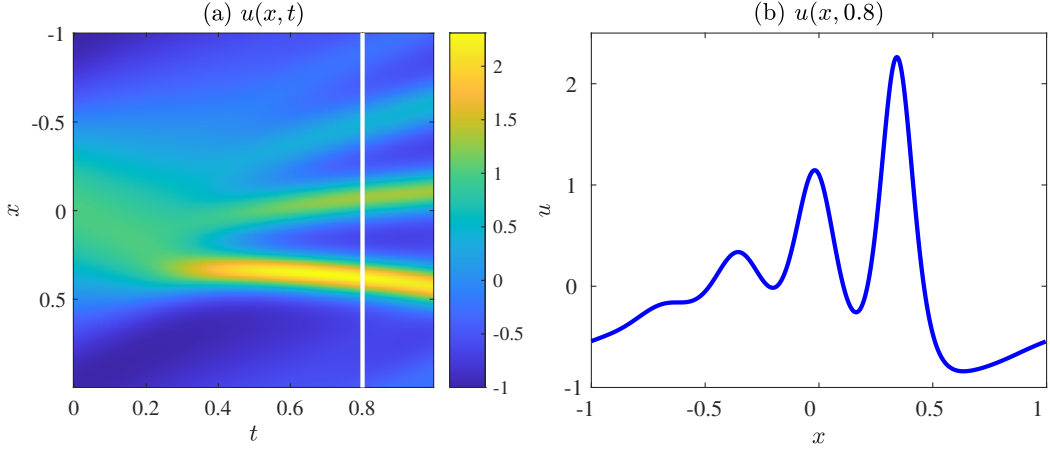}
	\caption{The traveling waves characterized by the KdV equation ($u_t = -uu_x-0.0025u_{xxx}$) with the initial condition $u(x,0)=\mathrm{cos}(\pi x)$ and periodic boundary conditions. $x\in[-1,1]$ and $t\in[0,1]$.}
	\label{Figure:KdV}
\end{figure}

The second example of learning PDE from a 1D dynamical system examines the effectiveness of the \textsc{PeSBL} method in correctly identifying governing equations containing higher-order spatial derivatives. This section considers a mathematical model of traveling waves on shallow water surfaces, i.e., the KdV equation with the form $u_t  =  \xi_1uu_x + \xi_2u_{xxx}$. The KdV equation can be used to characterize the evolution of many long 1D waves such as the ion acoustic waves in a plasma and acoustic waves on a crystal lattice \cite{raissi2019physics}. This study investigates the system described by the following KdV equation: $u_t = -uu_x-0.0025u_{xxx}$ with the initial condition $u(x,0)=\mathrm{cos}(\pi x)$ and periodic boundary conditions. Figure \ref{Figure:KdV} visualizes this system within the range $x\in[-1,1]$ and $t\in[0,1]$.

\begin{figure}[!h]
	\centering
	\includegraphics[scale=0.8]{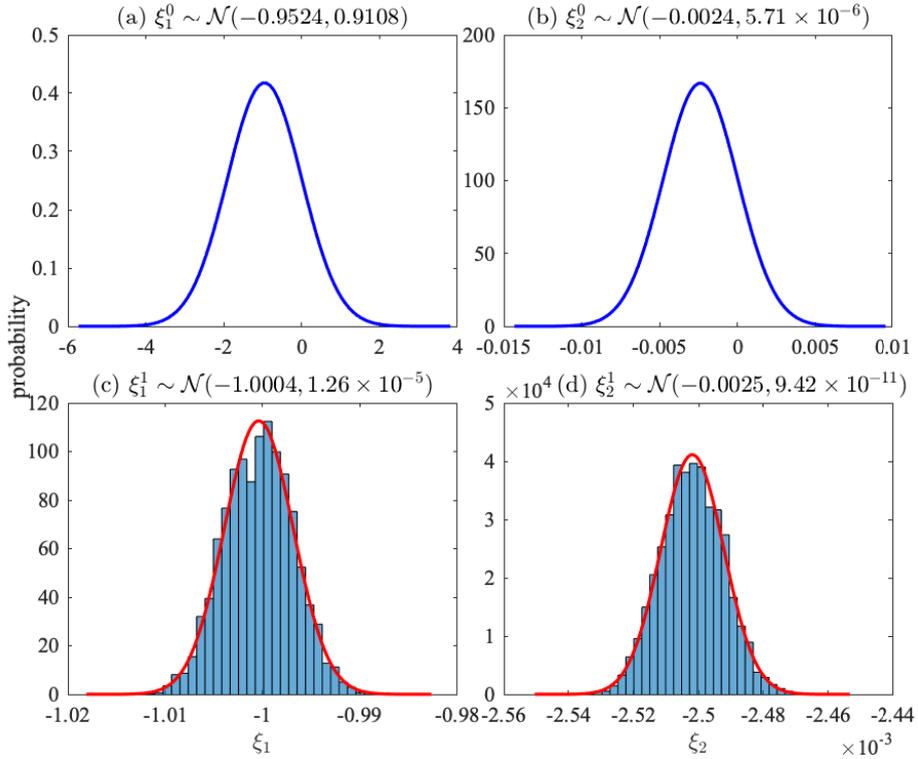}
	\caption{Results of sparse regression and BMU for the KdV equation ($u_t = -uu_x-0.0025u_{xxx}$) with 20\% noise. $\xi_1$ and $\xi_2$ denote the coefficient of $uu_x$ and $u_{xxx}$, respectively. The superscripts $^0$ and $^1$ denote the prior obtained from SBL and the posterior after BMU.}
	\label{Figure:BMU_KdVN20}
\end{figure}
 
\begin{table} [!h]
	\caption{Results of PDE learning using the PE-SBL method (KdV equation: $u_t = -uu_x-0.0025u_{xxx}$).}
	\begin{tabular}{ |c|c| } 
		\hline
		noise level &  identified PDE \\ \hline
		0\% & $u_t = -1.0000(\pm 0.0034)uu_x-0.0025(\pm 9.48\times 10^{-6})u_{xxx}$\\ \hline
		10\% &  $u_t = -1.0002(\pm 0.0035)uu_x-0.0025(\pm 9.40\times 10^{-6})u_{xxx}$\\ \hline
		20\% & $u_t = -1.0004(\pm 0.0035)uu_x-0.0025(\pm 9.70\times 10^{-6})u_{xxx}$\\ \hline
		50\% &  $u_t = -1.0012(\pm 0.0036)uu_x-0.0025(\pm 9.85\times 10^{-6})u_{xxx}$\\ \hline
	\end{tabular}\label{Table:kdvNoisy}
\end{table}

Table \ref{Table:kdvNoisy} summarizes the results of learning PDEs from the simulated traveling waves containing 0\% to 50\% noise. It shows that the correct PDE form with accurate coefficient means and limited uncertainties can be successfully identified using the \textsc{PeSBL} method in all simulated cases. Figure \ref{Figure:BMU_KdVN20} shows the results of PDE learning for this system with data containing 20\% measurement noise.

\subsection{Discovering PDEs for 2D systems}\label{results2D}
\begin{figure}[!h]
	\centering
	\includegraphics[scale=0.80]{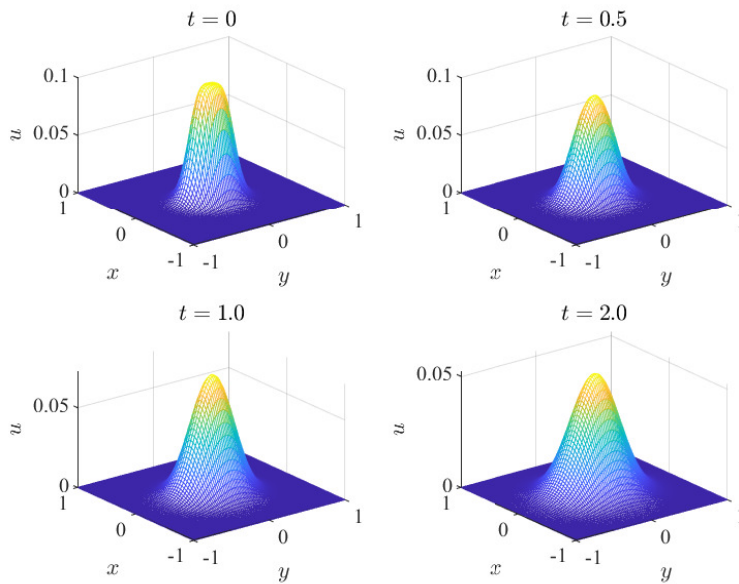}
	\caption{Dissipative system characterized by the 2D Burgers equation $u_t = -(uu_x+ uu_y)+0.01(u_{xx} +u_{yy})$.}
	\label{Figure:burgers2DN0}
\end{figure}

\begin{figure}[!h]
	\centering
	\includegraphics[scale=0.8]{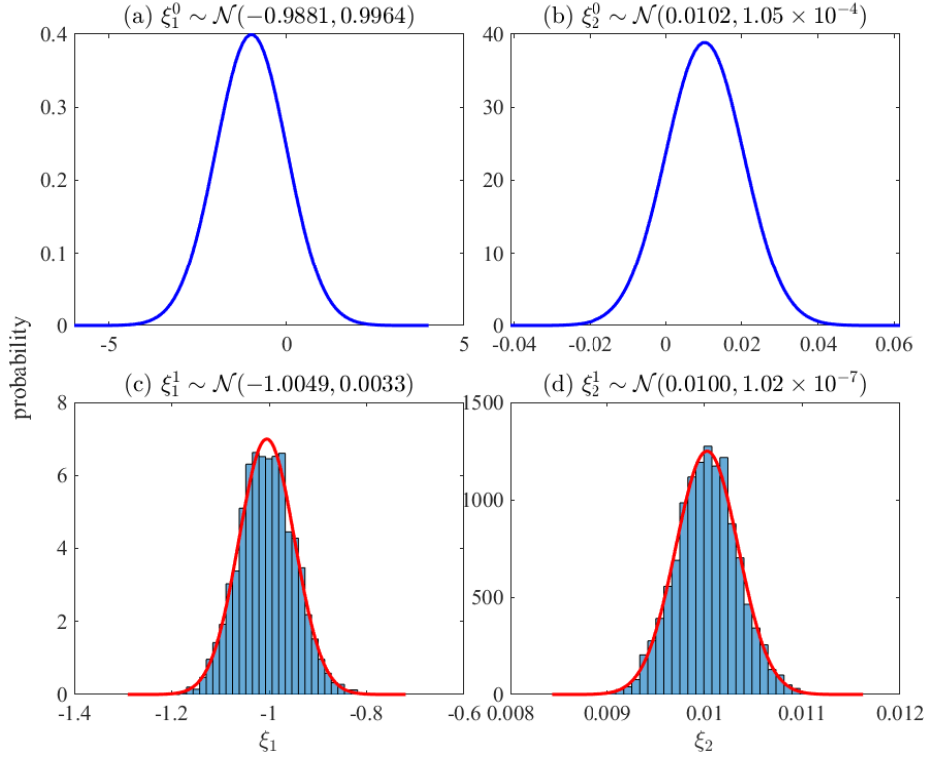}
	\caption{Results of sparse regression and BMU for 2D Burgers equation ( $u_t = -(uu_x+ uu_y)+0.01(u_{xx} +u_{yy})$) with 20\% noise. $\xi_1$ and $\xi_2$ denote the coefficient of $(uu_x+ uu_y)$ or ($\mathbf{u}\cdot\nabla)u$ and $(u_{xx} +u_{yy})$ or $\nabla^2u$, respectively. The superscripts $^0$ and $^1$ denote the prior obtained from SBL and the posterior after BMU.}
	\label{Figure:BMU_Burgers2DN20}
\end{figure}

This section investigates the effectiveness of the \textsc{PeSBL} method in learning the correct governing equation(s) characterizing 2D dynamical systems. First, the \textsc{PeSBL} method is used to discover the physics of a 2D dissipative system characterized by the Burgers equation $u_t = -(uu_x+ uu_y)+0.01(u_{xx} +u_{yy})$ with the initial condition $u(x,y,0) = 0.1\mathrm{sech}(20x^2+25y^2)$ and periodic boundary conditions. Figure \ref{Figure:burgers2DN0} shows the snapshots of simulated data for this system. It should be noted that with the increase of dimensionality, the knowledge discovery of dynamical systems becomes more complex. Hence, in the sparse regression scheme of PDE learning, the library matrix $\mathbf{\Theta}$ should no longer be built in an exhaustive manner considering all possible combinations of polynomials to a certain power and spatial derivatives to a certain order, which will make the sparse regression problem intractable. Instead, $\mathbf{\Theta}$ is built with representative terms in multi-dimensional nonlinear dynamical systems (e.g., convective derivative $\mathbf{u} \cdot \nabla$, advective acceleration $(\mathbf{u} \cdot 	\nabla)\mathbf{u}$, and the Laplacian $	\nabla^2(\mathbf{u})$) and their products with polynomials. Table \ref{Table:2dBurgers} summarizes the results of learning PDEs from the 2D dissipative system using simulated clean and noisy data. It shows that the correct equation form with accurate coefficient means and limited uncertainties can be successfully identified using the \textsc{PeSBL} method with data containing as much as 50\% random noise. Figure \ref{Figure:BMU_Burgers2DN20} shows the results of learning this 2D Burgers equation with data containing 20\% noise.

\begin{table} [!h]
	\caption{Results of PDE learning using the \textsc{PeSBL} method (2D Burgers equation: $u_t = -(uu_x+ uu_y)+0.01(u_{xx} +u_{yy})$).}
	\begin{tabular}{ |c|c| } 
		\hline
		noise level &  identified PDE \\ \hline
		0\% & $u_t = -0.9980(\pm 0.0564)(uu_x+ uu_y)+0.0100(\pm 3.16\times 10^{-4})(u_{xx} +u_{yy})$\\ \hline
		10\% &  $u_t = -1.0017(\pm 0.0556)(uu_x+ uu_y)+0.0100(\pm 3.18\times 10^{-4})(u_{xx} +u_{yy})$\\ \hline
		20\% & $u_t = -1.0049(\pm 0.0570)(uu_x+ uu_y)+0.0100(\pm 3.19\times 10^{-4})(u_{xx} +u_{yy})$\\ \hline
		50\% &  $u_t = -1.0153(\pm 0.0635)(uu_x+ uu_y)+0.0101(\pm 3.55\times 10^{-4})(u_{xx} +u_{yy})$\\ \hline
	\end{tabular}\label{Table:2dBurgers}
\end{table}

\begin{figure}[!h]
	\centering
	\includegraphics[scale=0.65]{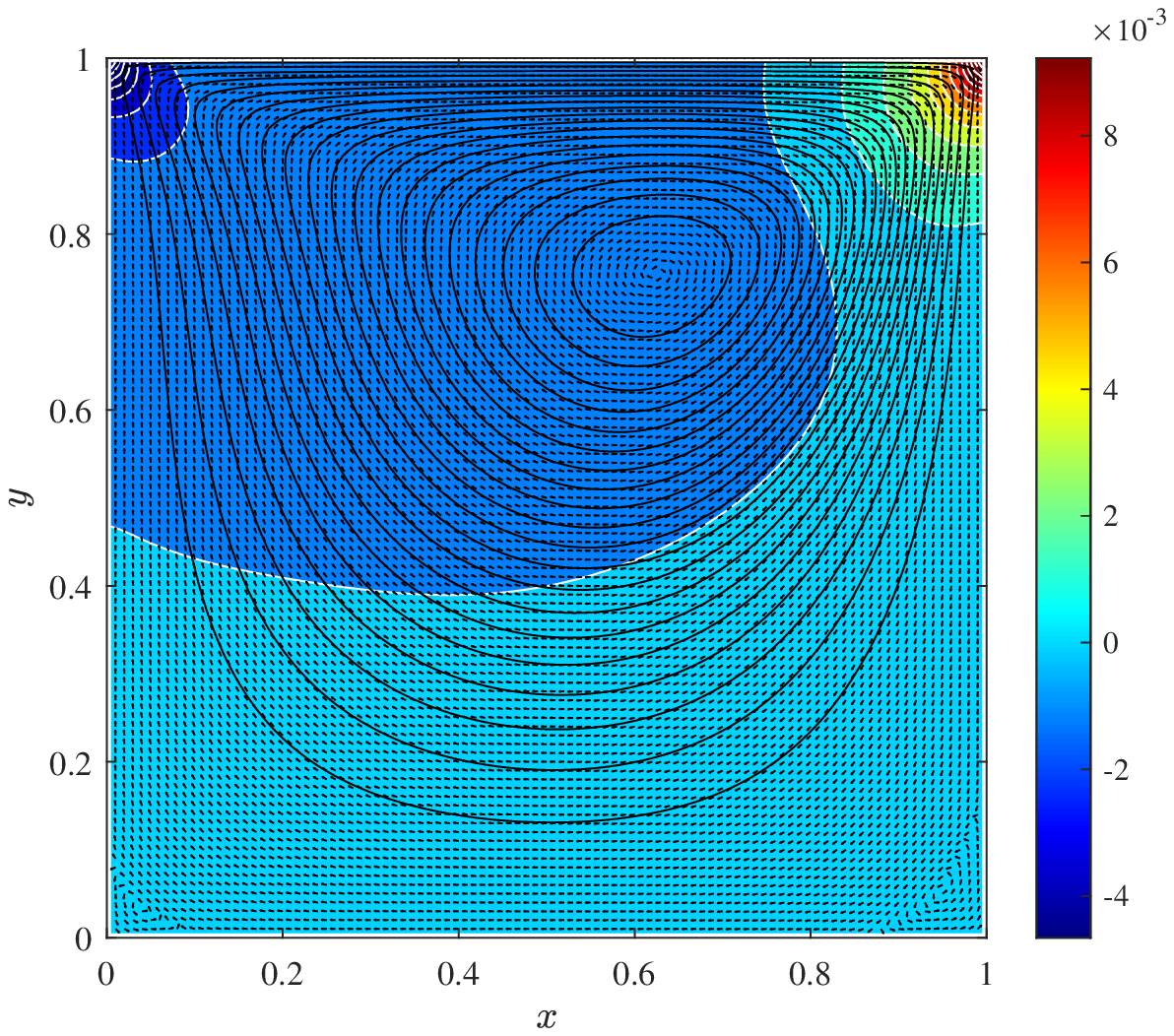}
	\caption{Simulated lid driven flow (at $t=4.0$ sec) characterized by the 2D Navier Stokes equation ( $\frac{\partial \mathbf{u}}{\partial t} = -(\mathbf{u}\cdot\nabla)\mathbf{u}-\nabla p + \frac{1}{100}\nabla^2\mathbf{u}$ with $\nabla\cdot\mathbf{u}=0$). The small arrows represent the velocity field; the color plot with contour line s denotes the pressure distribution; the closed contour lines are the streamlines.}
	\label{Figure:solNS_N0_2}
\end{figure}

Another 2D system investigated in this study is the lid-driven cavity flow which is a benchmark problem for viscous incompressible fluid flow \cite{Zienkiewicz2005finite}. This study uses a geometry of a square cavity that is comprised of a lid on the top moving with a tangential unit velocity and three no-slip rigid walls. The velocity and pressure distributions are numerically simulated for a Reynolds number of 100. Figure \ref{Figure:solNS_N0_2} visualizes this system at $t = 4.0$ sec, containing the velocity (small arrows) and pressure (color map with contour lines) distributions as well as the streamlines.

\begin{table} [!h]
	\caption{Results of PDE learning using the \textsc{PeSBL} method (2D Navier Stokes equation: $\frac{\partial \mathbf{u}}{\partial t} = -(\mathbf{u}\cdot\nabla)\mathbf{u}-\nabla p + \frac{1}{100}\nabla^2\mathbf{u}$ with $\nabla\cdot\mathbf{u}=0$).}
	\begin{tabular}{ |c|c| } 
		\hline
		noise level &  identified PDE \\ \hline
		0\% & $\frac{\partial \mathbf{u}}{\partial t} = -1.0000(\pm 0.0014)(\mathbf{u}\cdot\nabla)\mathbf{u}-1.0000(\pm 0.0014)\nabla p + \frac{1}{100.00}(\pm 1.38\times 10^{-5})\nabla^2\mathbf{u}$\\ \hline
		10\% & $\frac{\partial \mathbf{u}}{\partial t} = -1.0000(\pm 0.0013)(\mathbf{u}\cdot\nabla)\mathbf{u}-1.0000(\pm 0.0013)\nabla p + \frac{1}{100.00}(\pm 1.37\times 10^{-5})\nabla^2\mathbf{u}$ \\ \hline
		20\% & $\frac{\partial \mathbf{u}}{\partial t} = -1.0000(\pm 0.0014)(\mathbf{u}\cdot\nabla)\mathbf{u}-1.0001(\pm 0.0014)\nabla p + \frac{1}{99.99}(\pm 1.43\times 10^{-5})\nabla^2\mathbf{u}$\\ \hline
		50\% & $\frac{\partial \mathbf{u}}{\partial t} = -0.9999(\pm 0.0025)(\mathbf{u}\cdot\nabla)\mathbf{u}-1.0002(\pm 0.0023)\nabla p + \frac{1}{99.99}(\pm 2.29\times 10^{-5})\nabla^2\mathbf{u}$\\ \hline
	\end{tabular}\label{Table:NS}
\end{table}

\begin{figure}[!h]
	\centering
	\includegraphics[scale=0.7]{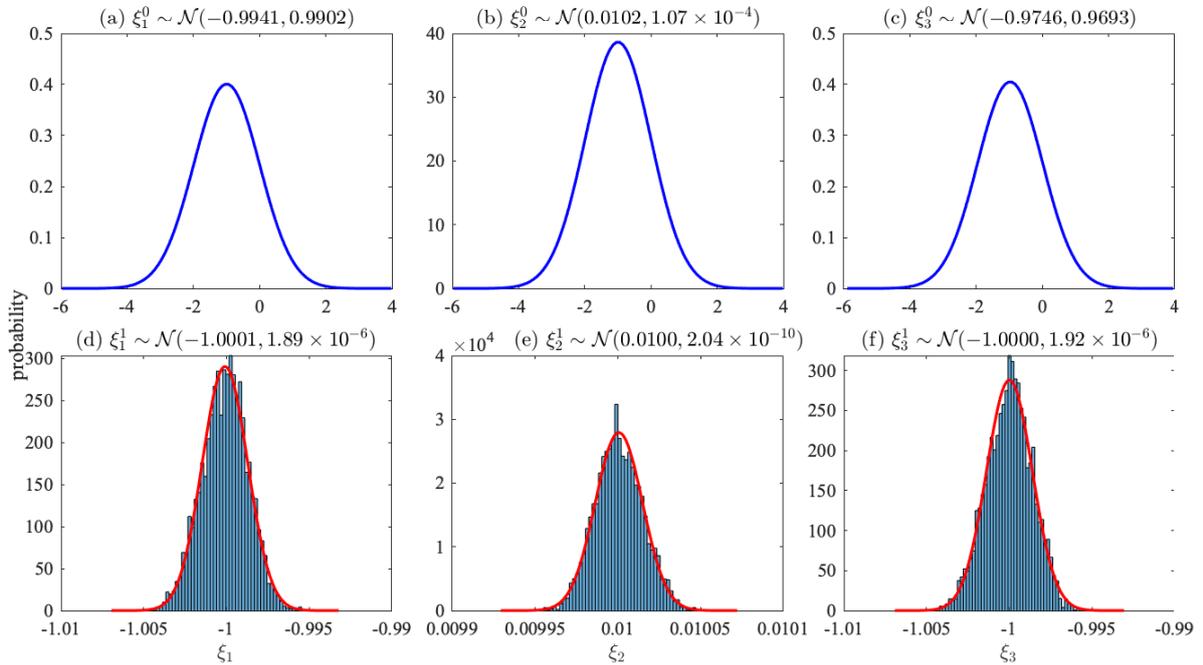}
	\caption{Results of sparse regression and BMU for 2D Navier Stokes equation ( $\frac{\partial \mathbf{u}}{\partial t} = -(\mathbf{u}\cdot\nabla)\mathbf{u}-\nabla p + \frac{1}{100}\nabla^2\mathbf{u}$ with $\nabla\cdot\mathbf{u}=0$) with 20\% noise. $\xi_1$ denotes the coefficient of $\nabla p$,  $\xi_2$ denotes the coefficient of $\nabla^2\mathbf{u}$, and $\xi_3$ denotes the coefficient of $(\mathbf{u}\cdot\nabla)\mathbf{u}$. The superscripts $^0$ and $^1$ denote the prior obtained from SBL and the posterior after BMU.}
	\label{Figure:BMU_NSN20}
\end{figure}

Table \ref{Table:NS} lists the results of PDE learning using the \textsc{PeSBL} method with simulated data containing various levels of noise. The correct PDE form can be learned for cases with as much as 50\% noise. When the noise level increases to 50\%, the standard deviations of model coefficients are almost doubled. This increased level of model uncertainty is mainly caused by the increase of noise level in addition to the challenge of learning PDE(s) for this complex system. However, this adverse influence is still well controlled mainly due to the efficient signal processing in the proposed method. Figure \ref{Figure:BMU_NSN20} shows the results of PDE learning for the case with 20\% measurement noise.

\subsection{Robustness of the \textsc{PeSBL} Method}
To highlight the merits of the \textsc{PeSBL} method in robustness, Table \ref{Table:burgersSBL_S3d} compares the results of learning the Burgers equation ($u_t = -uu_x+\frac{0.01}{\pi}u_{xx}$) using \textsc{PeSBL} method considering model parsimony and methods in the literature promoting only model sparsity. Without loss of generality, the simulated data with 20\% noise is used. It shows that the SBL method yields an as sparse model as the \textsc{PeSBL} method, however, with an incorrect and more complex diffusion term. This failure is due to that traditional RVM method (Section \ref{Section:RVM}) only promotes sparsity in regression and thus tends to include complex terms for marginal improvement in regression accuracy. More comparison about the learned form ($u_t = \xi_1uu_x+\xi_2u^2u_{xx}$) with the correct form ($u_t = \xi_1uu_x+\xi_2u_{xx}$) can be found the in the authors' previous study \cite{zhang2021robust}. The $\mathrm{S}^3\mathrm{d}$ method yields a complex PDE form which represents a totally intractable dynamical system.

\begin{table} [!h]
	\caption{Comparison of PDE learning results using methods with and without parsimony enhancement. The Burgers equation ($u_t = -uu_x+\frac{0.01}{\pi}u_{xx}$) is used for demonstration. The noise level in the measured data is 20\%.}
	\begin{tabular}{ |m{4.5cm}|m{10cm}| } 
		\hline
		method &  identified PDE \\ \hline
		\textsc{Pe}SBL (with parsimony enhancement)& $u_t = -1.0010(\pm0.0252)uu_x+\frac{0.0108}{\pi}(\pm 8.18\times10^{-4})u_{xx}$ \\ \hline
		SBL (without parsimony enhancement) & $u_t = -0.9773(\pm 0.9797)uu_x+0.0073(\pm 0.0074)u^2u_{xx}$\\ \hline
		$\mathrm{S}^3\mathrm{d}$ (without parsimony enhancement) \cite{yuan2019machine}& $u_t = -0.01381-0.1832u-0.1285u^3+0.0147u_x+0.0067u_{xx}+1.8\times 10^{-5}u_{xxx}-1.1525uu_x-0.0009uu_{xx}-0.0013uu_{xxx}-0.0039u^2u_{xx}+1.0385u^3u_{x}+0.0009u^3u_{xx}+0.0013u^3u_{xxx}$\\ \hline
	\end{tabular}\label{Table:burgersSBL_S3d}
\end{table}

\begin{table} [!h]
	\caption{Comparison of PDE learning results with and without signal preprocessing. The Burgers equation ($u_t = -uu_x+\frac{0.01}{\pi}u_{xx}$) is used for demonstration. The noise level in the measured data is 20\%.}
	\begin{tabular}{ |m{3.0cm}|m{10cm}| } 
		\hline
		method &  identified PDE \\ \hline
		\textsc{Pe}SBL without preprocessing& $u_t =-0.3296u+0.1844u^3-0.6474uu_x+0.3833u^3u_x-0.0059u_{xx}-0.0034u2^u_{xx}-0.0002uu_{xxx}+0.0001u^3u_{xxx}$ \\ \hline
		\textsc{Pe}SBL with preprocessing& $u_t = -1.0010(\pm0.0252)uu_x+\frac{0.0108}{\pi}(\pm 8.18\times10^{-4})u_{xx}$ \\ \hline
		SBL without preprocessing&$u_t =-0.3296u+0.1844u^3-0.6474uu_x+0.3833u^3u_x-0.0059u_{xx}-0.0034u^2u_{xx}-0.0002uu_{xxx}+0.0001u^3u_{xxx}$ \\ \hline
		SBL with preprocessing& $u_t = -0.9773(\pm 0.9797)uu_x+0.0073(\pm 0.0074)u^2u_{xx}$\\ \hline
		\end{tabular}\label{Table:burgersSBL_S3d2}
\end{table}

Additionally, Table \ref{Table:burgersSBL_S3d2} compares the results of PDE learning using the \textsc{PeSBL} and SBL methods with and without signal preprocessing. It shows that, without preprocessing the measured signals, the  \textsc{PeSBL} method cannot successfully identify the correct PDE form due to the significant influence from the measurement noise. With signal preprocessing, the  \textsc{PeSBL} method yields the correct model. This finding further approves the effectiveness of the signal preprocessing strategy established in this study (Section \ref{sec:preprocess}). The identical PDE form is learned through SBL using the measured raw data. When preprocessing is implemented, SBL yields a sparse equation form. Though with incorrect term, this equation is more tractable and has been proved to be considerably representative of the dissipative system in the authors' previous study \cite{zhang2021robust}. Therefore, the signal preprocessing strategy established in this study has the potential of improving the performance of other methods for PDE learning without parsimony enhancement.

\section{Propagating Model Uncertainties to System Dynamics} \label{uncertainProp}
Given noisy measurements from a certain dynamical system, the \textsc{PeSBL} method outputs the most probable model (i.e., PDE(s)) reflecting the underlying physics as well as the model uncertainties. The learned model uncertainties can be propagated to the simulated system dynamics, so that a confidence interval can be provided for the system responses at certain temporal and spatial coordinates. In this section, the two systems characterized by the Burgers equation ($u_t = -uu_x+\frac{0.01}{\pi}u_{xx}$) and the KdV equation ($u_t = -uu_x-0.0025u_{xxx}$) are used as examples to demonstrate the results of uncertainty propagation. Without loss of generality, the learned models together with their uncertainties in cases with 20\% measurement noise are used in this section. The results of uncertainty propagation for the 2D systems investigated in Section \ref{results2D} are not demonstrated in this section due to the expensive computational costs and difficulty in visualization. The codes for uncertainty propagation of all 1D and 2D systems investigated in Section \ref{Sec:results} can be found on the website: \href{https://github.com/ymlasu}{\textcolor{blue}{https://github.com/ymlasu}}.

\begin{figure}[!h]
	\centering
	\includegraphics[scale=0.75]{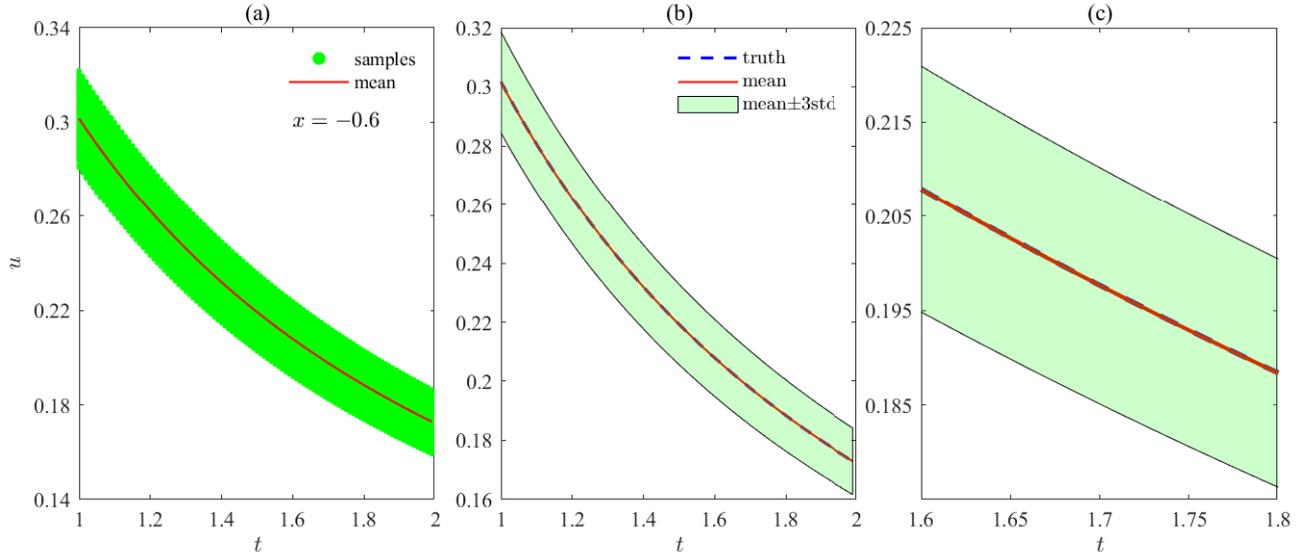}
	\caption{Estimated system response at $x = -0.6$ during the time interval ($1.0\leqslant t \leqslant 2.0$) for the system characterized by the Burgers equation ($u_t = -uu_x+\frac{0.01}{\pi}u_{xx}$). (a) simulated samples and the mean value; (b) comparison of the sample mean and standard deviation with the truth; (c) an amplified view of (b) in the range ($1.6\leqslant t \leqslant 1.8$).}
	\label{Figure:BurgersUP_N20}
\end{figure}

Unlike linear systems for which the model parameters and its outputs are explicitly related, the nonlinear dynamical system has an implicit relationship between the model parameters and outputs. Therefore, for a certain system in this study, the model uncertainty is propagated to the system responses by numerically simulating the system with the rich parameter samples obtained from MCMC in BMU. Summarizing the results of numerical simulations, the uncertainties of system dyanmics can be quantified and analyzed. To effectively demonstrate the representativeness of learned models and the efficiency of uncertainty quantification and propagation, the responses of the two systems at ``future'' time ($t\geqslant 1$) are simulated and presented in this section. 

\begin{figure}[!h]
	\centering
	\includegraphics[scale=0.75]{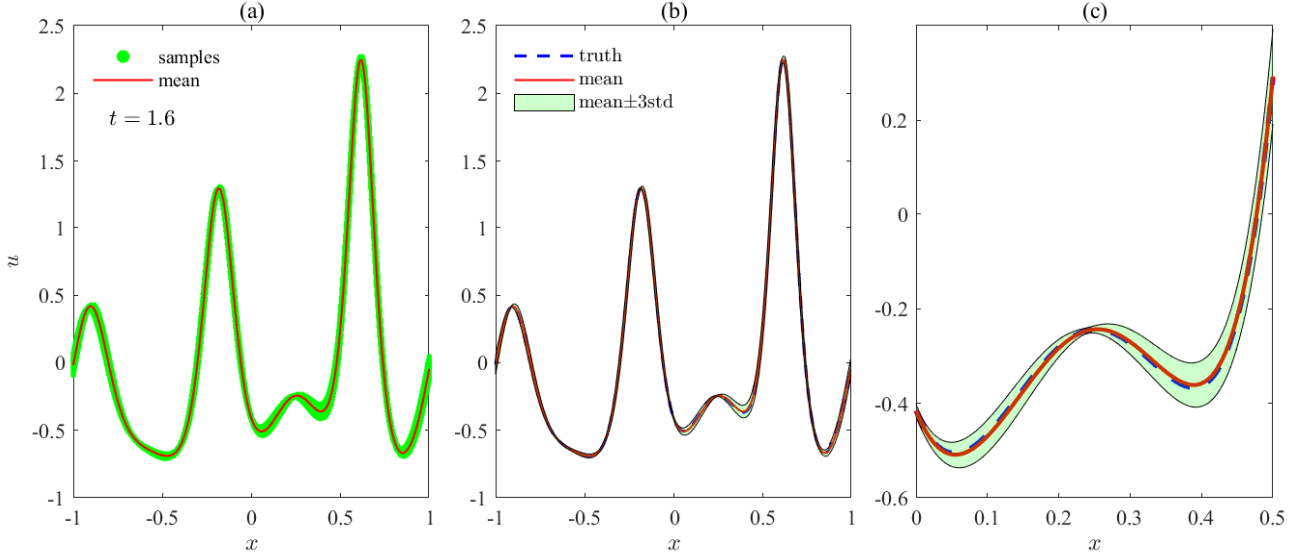}
	\caption{Estimated system response at $t = 1.6$ for the system characterized by the KdV equation ($u_t = -uu_x-0.0025u_{xxx}$). (a) simulated samples and the mean value; (b) comparison of the sample mean and standard deviation with the truth; (c) an amplified view of (b) in the range ($0\leqslant x \leqslant 0.5$).}
	\label{Figure:KdVUP_N20}
\end{figure}

Figure \ref{Figure:BurgersUP_N20} (a) plots the simulated samples at the cross section $x=-0.6$ and $1\leqslant t \leqslant 2$ and their mean value for the dissipative system characterized by the Burgers equation. It can be observed that the system uncertainty remains at nearly the same level within this region, most probably due to the smoothness of the system therein and the constant uncertainty from the learned model. Figure \ref{Figure:BurgersUP_N20} (b) (amplified in (c)) compares the sample mean with the underlying truth of the system at the same cross section. It shows that the mean of samples deviates slightly from the true value which is always encompassed within the range (mean-3std,mean+3std) (std stands for standard deviation). This observation approves the reliability of the learned model using the proposed \textsc{PeSBL} method in accurately predicting systems responses. Figures \ref{Figure:KdVUP_N20} (a) to (c) show the results of uncertainty propagation for the traveling waves characterized by the KdV equation, the system responses at $t = 1.6$ are plotted for demonstration. Due to the sharp transition of the system responses at this cross section, the propagated uncertainty varies in magnitude as $x$ increases from -1 to 1. Nevertheless, the mean of samples stays close to the true value which is always covered by the statistic interval (mean-3std,mean+3std). 

\section{System Diagnosis and Prognosis}\label{Section:DAP}

\begin{figure}[!h]
	\centering
	\includegraphics[scale=0.8]{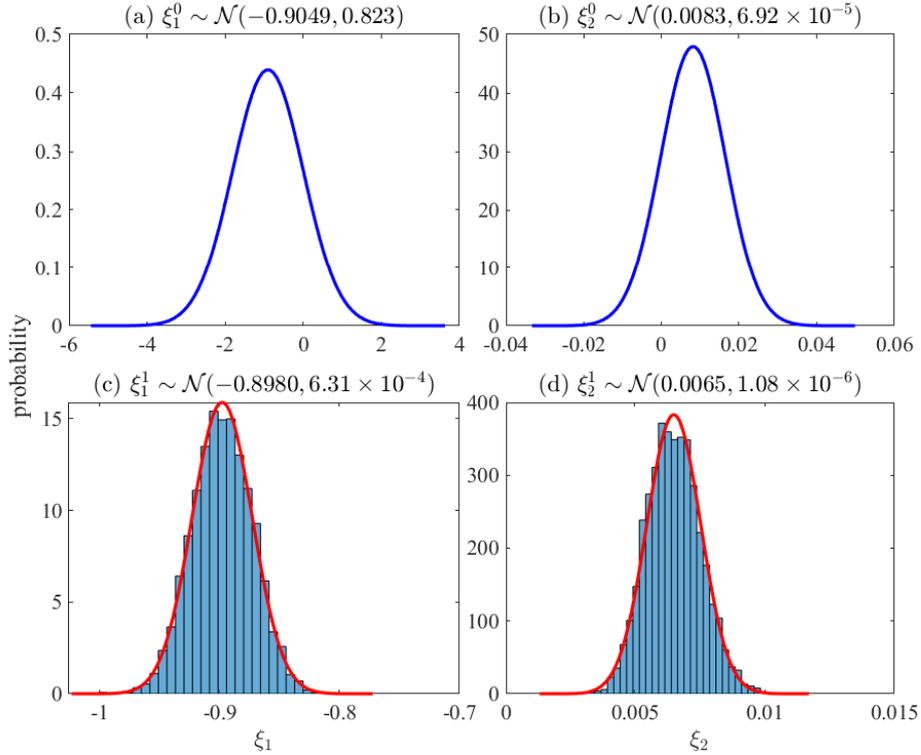}
	\caption{Results of sparse regression and BMU for the Burgers equation ($u_t = -0.9uu_x+\frac{0.02}{\pi}u_{xx}$) with 20\% noise. Please refer to the caption of Figure \ref{Figure:BMU_burgersN0} for more information.}
	\label{Figure:BMU_burgers1N20}
\end{figure}

\begin{figure}[!h]
	\centering
	\includegraphics[scale=0.8]{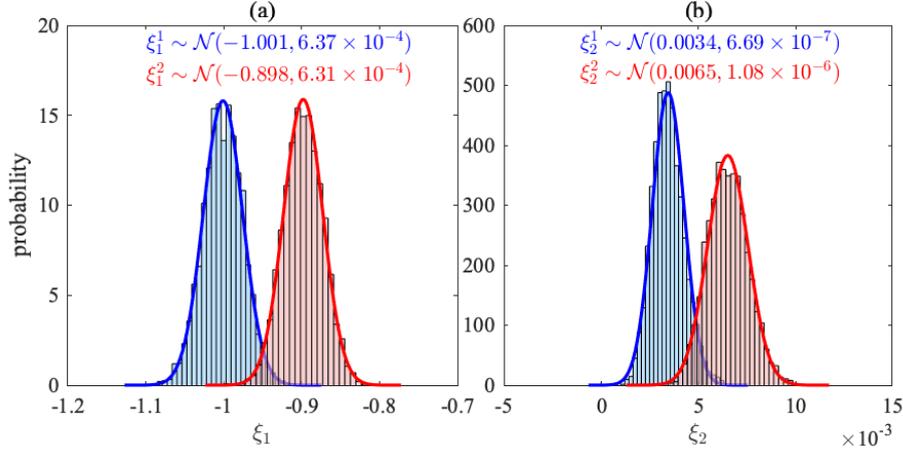}
	\caption{Comparison of BMU results for two dissipative systems characterized by the Burgers equation ($u_t = -uu_x+\frac{0.01}{\pi}u_{xx}$ and $u_t = -0.9uu_x+\frac{0.02}{\pi}u_{xx}$) with data containing 20\% noise. In this figure, the superscript $^1$ denotes the results from system 1 characterized by Burgers equation $u_t = -uu_x+\frac{0.01}{\pi}u_{xx}$,  and the superscript $^2$ denotes the results from system 2 characterized by Burgers equation $u_t = -0.9uu_x+\frac{0.02}{\pi}u_{xx}$. Please refer to the caption of Figure \ref{Figure:BMU_burgersN0} for more information.}
	\label{Figure:BMU_Burgers_diagnosis1_N20}
\end{figure}

This section examines the capability of the \textsc{PeSBL} framework in system diagnosis when significant variation occurs to the investigated system. The dissipative system characterized by the Burgers equation ($u_t = \xi_1uu_x+\xi_2u_{xx}$) is used as an example in this analysis without loss of generality. In this example analysis, the simulated data contains 20\% noise. To simulate the variation of this system, both advection and diffusion coefficients are varied such that the governing equation changes from $u_t = -uu_x+\frac{0.01}{\pi}u_{xx}$ to $u_t = -0.9uu_x+\frac{0.02}{\pi}u_{xx}$. Figure \ref{Figure:BMU_burgers1N20} shows the results of sparse regression and BMU using the measured data from the varied system. The learned stochastic PDE has the form $u_t = -0.8980(\pm0.0251)uu_x+\frac{0.0204}{\pi}(\pm 0.001)u_{xx}$, which significantly deviates from that of the original system (i.e., $u_t = -1.0010(\pm0.0252)uu_x+\frac{0.0108}{\pi}(\pm 8.18\times10^{-4})u_{xx}$). Figures \ref{Figure:BMU_Burgers_diagnosis1_N20} (a) and (b) compare the posterior PDFs of advection and diffusion coefficients (i.e., $\xi_1$ and $ \xi_2$ respectively) of the original and varied systems. It can be observed that the governing model of the investigated system has shifted considerably with a high probability. These outcomes with uncertainties enable us to diagnose possible change of the system in a probabilistic manner.  For example, with the fitted normal posteriors shown in the figure, the variation of the advection and diffusion coefficients can be evaluated as $\delta\xi_1 \sim \mathcal{N}(0.103,0.0013)$ and $\delta\xi_2 \sim \mathcal{N}(0.0031,1.75\times 10^{-6})$, respectively. It remains to be examined whether this framework can be used for the prognostic health management of real physical systems. 

\section{Multiscale Modeling through Hierarchical Bayesian Inference (HBI)}\label{Section:MSM}

In Sections \ref{Sec:method} to \ref{Section:DAP}, the PDE learning is implemented with a certain dataset measured from a certain system, and the learned model uncertainty comes from the disturbance of other candidate terms in the library $\boldsymbol{\Theta}$ and the influence of measurement noise. However, in reality, the system may vary from time to time due to the change of environmental and other factors (such as temperature change and external excitations). This variation of system makes the corresponding model intrinsically uncertain, and this intrinsic model uncertainty cannot be quantified using the Bayesian inference framework established in Section \ref{Sec:BMU}. Therefore, the multiscale Bayesian modeling of dynamical systems is investigated in this section through Hierarchical Bayesian Inference (HBI) to quantify the intrinsic model uncertainty of systems. 
\subsection{Framework}\label{Sec:HBI-frame}
\begin{figure*}[!h]
\centering
\includegraphics[scale=1.0]{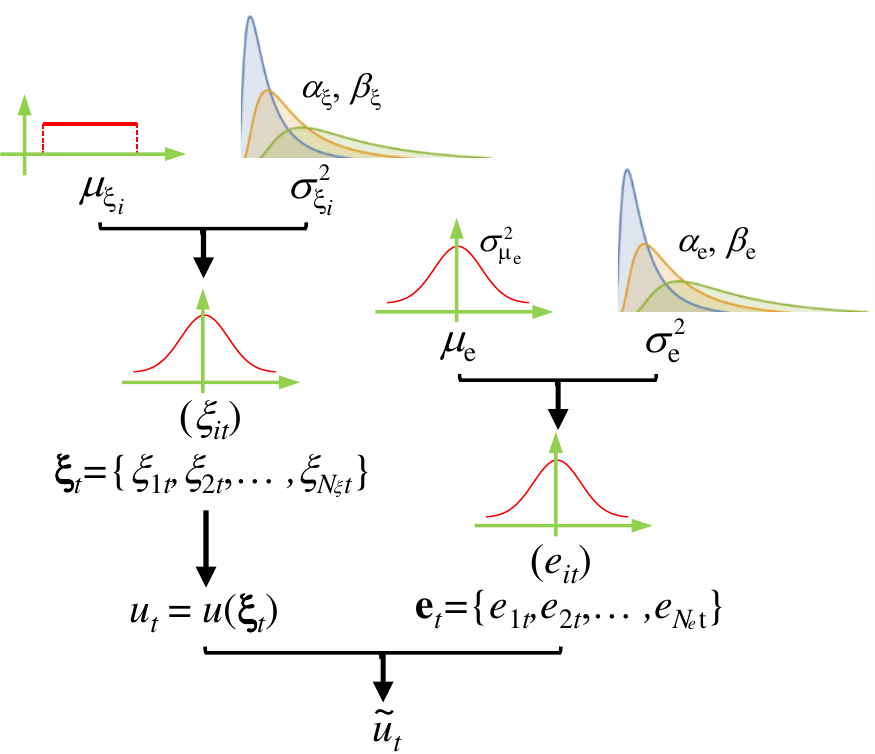}
\caption{Graphical model for hierarchical Bayesian modeling.}\label{Figure:HBI}
\end{figure*}

Figure \ref{Figure:HBI} illustrates the graphical HBI model for multiscale modeling of dynamical systems. It explicitly describes the relationship between measurements from a certain system and its model parameters and hyperparameters, and thus helps deriving the posterior distribution and full conditional distributions. In this framework, the system model during a certain test $t$ has model parameters $\boldsymbol{\xi}_t$ which is a sample of its underlying distribution. In this study, it is assumed that $\xi_{it}$ follows a Gaussian distribution such that $\xi_{it} \sim \mathcal{N}\left(\mu_{\xi_i},\sigma_{\xi_i}^2\right)$. Given the measurement during a certain test $\widetilde{u}_t$, the error function in this test can be established in a similar way to Equations \ref{Eq:err} and \ref{Eq:err2}, such that

\begin{equation}
\boldsymbol{e}_t = \frac{\widetilde{u}_t-u(\boldsymbol{\xi}_t)}{  \lVert \widetilde{u}_t \rVert}
\end{equation}
with

\begin{equation}
\mathrm{i.i.d.} \quad e_{it} \sim \mathcal{N}(\mu_e,\sigma_e^2)
\end{equation}
in which $t=1,2,...N_t$ and $N_t$ is the number of tests conducted. Following Equation \ref{Eq:llh}, the likelihood function can be built as
\begin{equation}\label{Eq:llh2}
\begin{split}
p(\widetilde{\boldsymbol{u}} \big\lvert \boldsymbol{\Xi},\mu_e,\sigma_e^2)&= \prod_{t=1}^{N_t}p(\widetilde{u}_t\big\lvert \boldsymbol{\xi}_t,\mu_e,\sigma_e^2) \\
&= \prod_{t=1}^{N_t}p(\boldsymbol{e}_t\big\lvert \mu_e,\sigma_e^2) \\
&= \prod_{t=1}^{N_t}\prod_{i=1}^{N_e}\mathcal{N}(e_{it}\big\lvert \mu_e,\sigma_e^2)
\end{split}
\end{equation}
in which $\boldsymbol{\Xi}=\left\lbrace \boldsymbol{\xi}_1, \boldsymbol{\xi}_2, ..., \boldsymbol{\xi}_{N_t} \right\rbrace$ is the collection of model parameters in all tests. The prior of $\mu_{\xi_i}$ is designed as a uniform distribution with upper limit $\mu_{\xi_i}^\mathrm{l}$ and lower limit $\mu_{\xi_i}^\mathrm{u}$ as shown in Equation \ref{Eq:prior1_2}. These limits are estimated from literature survey or prior knowledge of experts. 
\begin{equation}\label{Eq:prior1_2}
\mu_{\xi_i}\sim U(\mu_{\xi_i}^\mathrm{l},\mu_{\xi_i}^\mathrm{u})
\end{equation}
Without much knowledge about the uncertainties of the model parameters, their variance are assumed following the same non-informative inverse gamma prior, such that
\begin{equation}
\sigma_{\xi_i}^2 \sim Inv\-Gamma(\alpha_\xi,\beta_\xi)
\end{equation}
in which $\alpha_\xi=1$ and $\beta_\xi=2$. The same priors for $\mu_e$ and $\sigma_e$ as that in Equations \ref{Eq:prior2} and {Eq:prior3} are used and rewritten as follows:
\begin{equation}
\mu_e \sim \mathcal{N}(0,\sigma_{\mu_e}^2)
\end{equation}
\begin{equation}\label{Eq:prior2_2}
\sigma_e^2 \sim Inv\-Gamma(\alpha_e,\beta_e)
\end{equation}

With priors and the likelihood function defined, the posterior PDF of the parameter set can be derived as:
\begin{equation}\label{eq:pos_2}
\begin{split}
&p\left(
\boldsymbol{\Xi},\boldsymbol{\mu}_\xi,
\boldsymbol{\Sigma}_\xi,
\mu_e,\sigma_e
\big\lvert
\widetilde{\boldsymbol{u}}
\right)
\propto \\
&\prod_{t=1}^{N_t}
p\left(\widetilde{u}_t\big\lvert\boldsymbol{\xi}_t, \mu_e,\sigma_e^2\right)
p\left( \boldsymbol{\xi}_t\big\lvert\boldsymbol{\mu}_\xi, \boldsymbol{\Sigma}_\xi\right)
p\left(\boldsymbol{\mu}_\xi\right) p\left(\boldsymbol{\Sigma}_\xi\right)
p\left(\mu_e\right) p\left(\sigma_e^2\right)
\end{split}
\end{equation}
in which $\boldsymbol{\mu}_\xi=\left\lbrace  \mu_{\xi_1},\mu_{\xi_2},..., \mu_{\xi_{N_\xi}} \right\rbrace$ contains the mean of each model parameter, $\boldsymbol{\Sigma}_\xi$ is the covariance matrix of all model parameters and is a diagonal matrix with $\boldsymbol{\Sigma}_\xi(i,i) = \sigma_{\xi_i}^2$ $(i=1,2,...,N_\xi)$, and $\widetilde{\boldsymbol{u}}=\left\lbrace \widetilde{u}_1, \widetilde{u}_2, ..., \widetilde{u}_{N_t}\right\rbrace$ is the collection of measurement in all tests. Plugging in the expressions of priors and likelihood function, the posterior PDF becomes

\begin{equation}\label{Eq:postPDF}
\begin{split}
&p\left(\boldsymbol{\xi},\mu_e,\sigma_e^2 \big\lvert \widetilde{u}\right)
\propto\\
&\prod_{t=1}^{N_t}\left\lbrace\prod_{i=1}^{N_e}\frac{1}{\sqrt{\sigma_e^2}}\mathrm{exp}\left[-\frac{1}{2}\left(\frac{e_{it}-\mu_e}{\sigma_e}\right)^2\right]\right\rbrace
\left\lbrace\prod_{i=1}^{N_\xi}\frac{1}{\sqrt{\sigma_{\xi_i}^2}}\mathrm{exp}\left[-\frac{1}{2}\left(\frac{\xi_{it}-\mu_{\xi_i}}{\sigma_{\xi_i}}\right)^2\right]\right\rbrace\\
& \prod_{i=1}^{N_\xi}1\left(\mu_{\xi_{i}}^l<\mu_{\xi_{i}}<\mu_{\xi_{i}}^u\right)
\left[\prod_{i=1}^{N_\xi}\left(\sigma_{\xi_i}^2\right)^{-\alpha_\xi-1}\mathrm{exp}\left(-\frac{\beta_\xi}{\sigma_{\xi_i}^2}\right)\right]\\
&\mathrm{exp}\left[-\frac{1}{2}\left(\frac{\mu_e}{\sigma_{\mu_e}}\right)^2\right]
(\sigma_e^2)^{-\alpha_e-1}\mathrm{exp}\left(-\frac{\beta_e}{\sigma_e^2}\right)\\
&\propto (\sigma_e^2)^{-\frac{N_tN_e}{2}-\alpha_e-1}\prod_{i=1}^{N_\xi}(\sigma_{\xi_i}^2)^{-\frac{N_t}{2}-\alpha_\xi-1}
\times1\left(\mu_{\xi_{i}}^l<\mu_{\xi_{i}}<\mu_{\xi_{i}}^u\right)\\
&\mathrm{exp}\left[-\frac{1}{2}\sum_{t=1}^{N_t}\sum_{i=1}^{N_e}\left(\frac{e_{it}-\mu_e}{\sigma_e}\right)^2
-\frac{1}{2}\sum_{t=1}^{N_t}\sum_{i=1}^{N_\xi}\left(\frac{\xi_{it}-\mu_{\xi_i}}{\sigma_{\xi_i}}\right)^2
-\sum_{i=1}^{N_\xi}\frac{\beta_\xi}{\sigma_{\xi_i}^2}
-\frac{1}{2}\left(\frac{\mu_e}{\sigma_{\mu_e}}\right)^2
-\frac{\beta_e}{\sigma_e^2}\right]
\end{split}
\end{equation}

\begin{figure}[!h]
	\centering
	\includegraphics[scale=0.8]{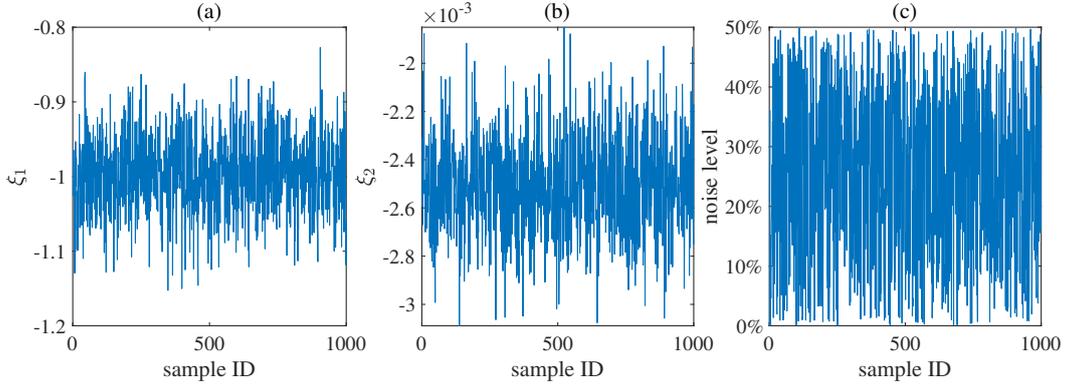}
	\caption{Model and noise parameters generated for the KdV equation ($u_t = \xi_1uu_x+\xi_2u_{xxx}$). (a) $\xi_1$: the coefficient of $uu_x$; (b) $\xi_2$: the coefficient of $u_{xxx}$ (c) the noise level in each dataset.}
	\label{Figure:dataHBI}
\end{figure}

Following the principle in Equation \ref{Eq:conditional}, the full conditional posterior distributions can be derived as follows:
\begin{equation} \label{eq:cod_pos1}
p\left(\xi_{it}\big\vert\cdot\right)
\propto
\mathrm{exp}\left[-\frac{1}{2}\sum_{i=1}^{N_e}\left(\frac{e_{it}-\mu_e}{\sigma_e}\right)^2
-\frac{1}{2}\left(\frac{\xi_{it}-\mu_{\xi_i}}{\sigma_{\xi_i}}\right)^2\right]
\end{equation}
\begin{equation}\label{Eq:trcN}
\begin{split}
\left(\mu_{\xi_i}\big\vert\cdot\right) &\propto
\mathrm{exp}\left[-\frac{1}{2}\sum_{t=1}^{N_t}\left(\frac{\xi_{it}-\mu_{\xi_i}}{\sigma_{\xi_i}}\right)^2\right]
\times1\left(\mu_{\xi_{i}}^l<\mu_{\xi_{i}}<\mu_{\xi_{i}}^u\right)\\
&\sim\mathcal{N}_{\mu_{\xi_i}^l}^{\mu_{\xi_i}^u}\left(\frac{1}{N_t}\sum_{t=1}^{N_t}\xi_{it},\frac{1}{N_t}\sigma_{\xi_i}^2\right)
\end{split}
\end{equation}

\begin{figure}[!htb]
	\centering
	\includegraphics[scale=0.8]{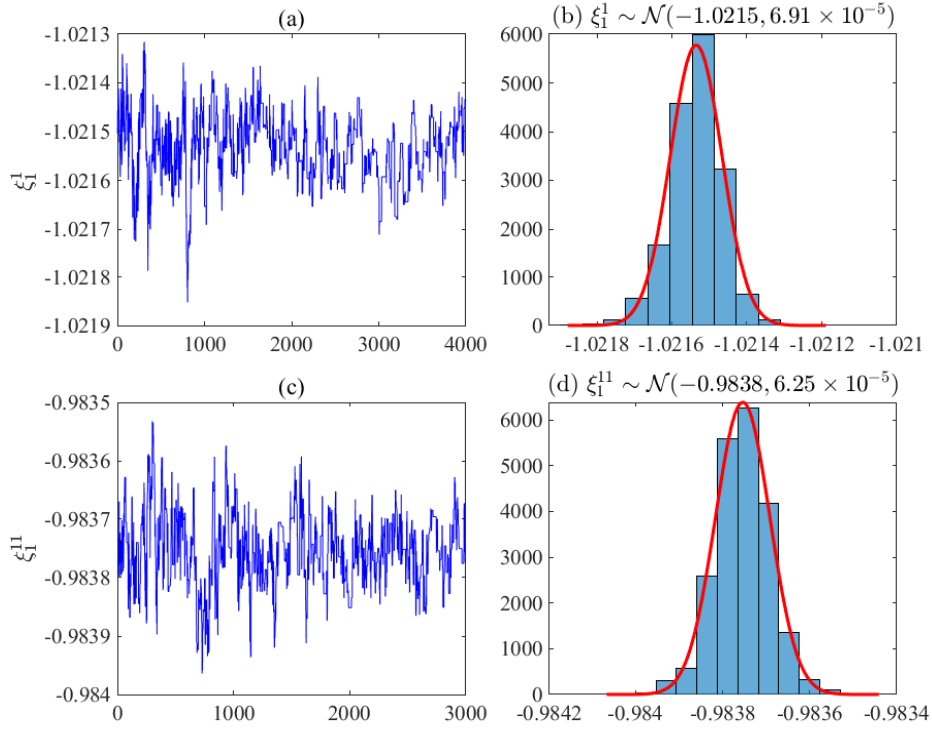}
	\caption{Example simulated samples and their distributions for $\xi_1^t$. Superscript $^t$ denotes the test number. (a) samples of $\xi_1^1$; (b) distribution of samples in (a) and the fitted Gaussian distribution; (c) samples of $\xi_1^{11}$; (d) distribution of samples in (c) and the fitted Gaussian distribution.}
	\label{Figure:HBI_KdV_xi1}
\end{figure}

\begin{equation}
\begin{split}
\left(\sigma_{\xi_i}^2\big\vert\cdot\right) 
&\propto
 (\sigma_{\xi_i}^2)^{-\frac{N_t}{2}-\alpha_\xi-1}
\mathrm{exp}\left[-\frac{1}{2}\sum_{t=1}^{N_t}\left(\frac{\xi_{it}-\mu_{\xi_i}}{\sigma_{\xi_i}}\right)^2
-\frac{\beta_\xi}{\sigma_{\xi_i}^2}\right]\\
&\sim Inv\-Gamma\left(\frac{N_t}{2}+\alpha_\xi,\frac{1}{2}\sum_{t=1}^{N_t}(\xi_{it}-\mu_{\xi_i})^2+\beta_\xi\right)\\
\end{split}
\end{equation}

\begin{equation} \label{eq:cod_pos2}
\begin{split}
p\left(\mu_e\big\vert\cdot\right)
&\propto
\mathrm{exp}\left[-\frac{1}{2}\sum_{t=1}^{N_t}\sum_{i=1}^{N_e}\left(\frac{e_{it}-\mu_e}{\sigma_e}\right)^2
-\frac{1}{2}\left(\frac{\mu_e}{\sigma_{\mu_e}}\right)^2\right]\\
&\sim\mathcal{N}\left(\frac{\sum_{t=1}^{N_t}\sum_{i=1}^{N_e}e_{it}}{N_tN_e+\frac{\sigma_e^2}{\sigma_{\mu_e}^2}},\frac{1}{\frac{N_tN_e}{\sigma_e^2}+\frac{1}{\sigma_{\mu_e}^2}}\right)
\end{split}
\end{equation}
\begin{equation}
\begin{split}
\left(\sigma_e^2\big\vert\cdot\right) 
&\propto
 (\sigma_e^2)^{-\frac{N_tN_e}{2}-\alpha_e-1}
\mathrm{exp}\left[-\frac{1}{2}\sum_{t=1}^{N_t}\sum_{i=1}^{N_e}\left(\frac{e_{it}-\mu_e}{\sigma_e}\right)^2
-\frac{\beta_e}{\sigma_e^2}\right]\\
&\sim Inv\-Gamma\left(\frac{N_tN_e}{2}+\alpha_e,\frac{1}{2}\sum_{t=1}^{N_t}\sum_{i=1}^{N_e}(e_{it}-\mu_e)^2+\beta_e\right)\\
\end{split}
\end{equation}

\subsection{A Case Study}
The system for traveling waves on shallow water surfaces characterized by the KdV equation ($u_t  =  \xi_1uu_x + \xi_2u_{xxx}$) is used for a case study in this section. 1000 models are randomly generated with $\xi_1 \sim \mathcal{N}(-1,0.05^2)$ and $\xi_2 \sim \mathcal{N}(-0.0025,0.0002^2)$, and one dataset is simulated for each model with the noise level randomly sampled from the uniform distribution $U(0,50\%)$. Figure \ref{Figure:dataHBI} shows generated samples of model parameters and noise level in the simulated data. Considering the computational cost in Bayesian inference via MCMC, 20 sets of data will be used for multiscale Bayesian modeling in this section using the HBI framework established in Section \ref{Sec:HBI-frame}. 

\begin{figure}[!h]
	\centering
	\includegraphics[scale=0.8]{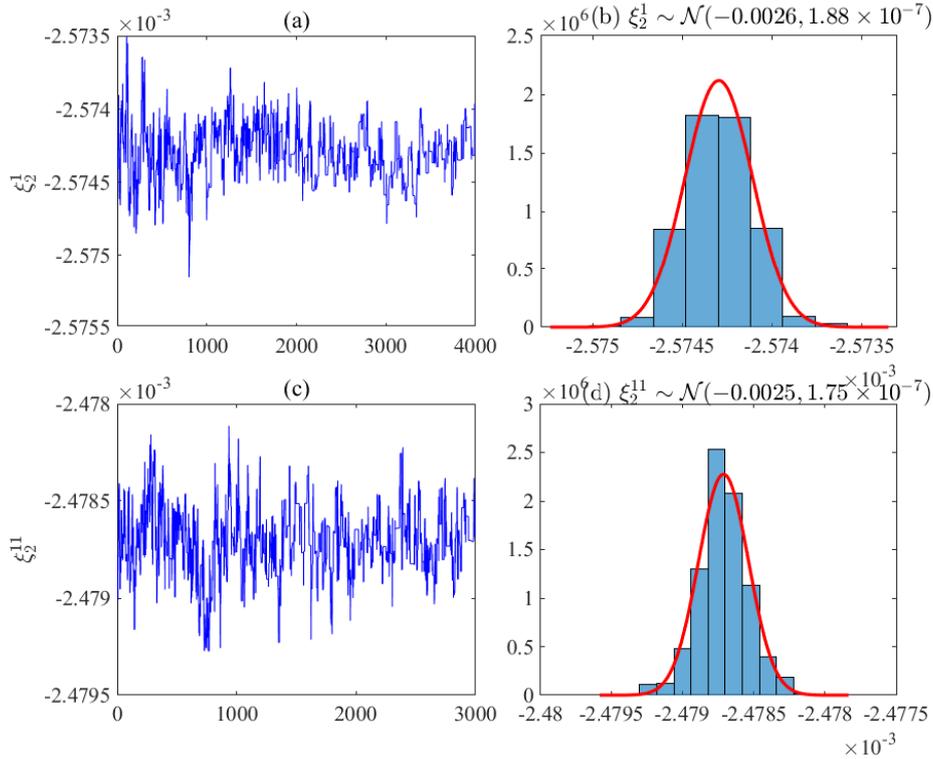}
	\caption{Example simulated samples and their distributions for $\xi_2^t$. Please refer to the caption of Figure \ref{Figure:HBI_KdV_xi1} for more information.}
	\label{Figure:HBI_KdV_xi2}
\end{figure}

\begin{figure}[!h]
	\centering
	\includegraphics[scale=0.8]{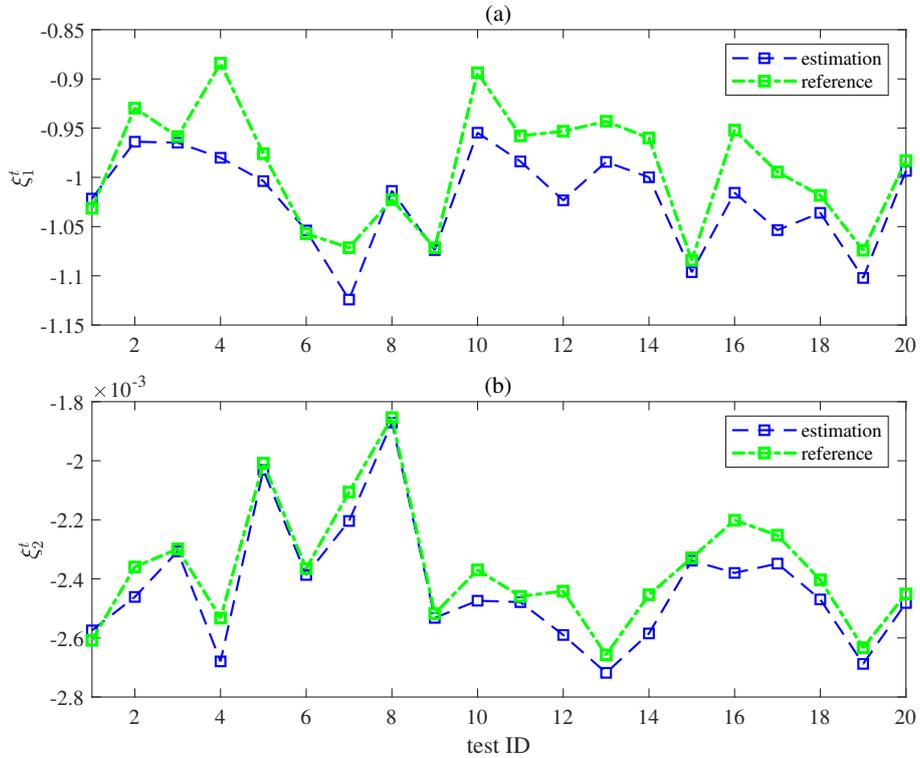}
	\caption{Comparison between the estimated model parameters via Bayesian inference and the reference values. (a) $\xi_1^t$; (b) $\xi_2^t$.}
	\label{Figure:HBI_KdV_xi_t}
\end{figure}

With the expressions of posteriors derived, the Bayesian inference problem for multiscale modeling is solved numerically via MCMC. 5000 samples are simulated for each model parameter. It may take more iterations for certain parameter to converge than others. Samples simulated prior to the convergence will be excluded for visualization and further analysis. Figures \ref{Figure:HBI_KdV_xi1} and \ref{Figure:HBI_KdV_xi2} show the simulated samples of model parameters $\xi_1$ and $\xi_2$ in two example tests and their distributions. A Gaussian distribution is fitted from the simulated samples. Its mean value is used as the estimated model parameter in a certain test, and its variance quantifies the uncertainty from measurement noise and numerical simulation. Figure \ref{Figure:HBI_KdV_xi_t} compare the estimated model parameters in the 20 tests with the reference values numerically generated in the beginning. Figure \ref{Figure:HBI_KdV_xi_t} (a) shows that the estimated values of $\xi_1^t$ remain close to the reference values in all tests with an average relative error of 3.49\%. All tests have a relative error below 10\% except test 4 in which the relative error is 10.86\%. Compared with $\xi_1^t$, the estimation of $\xi_2^t$ has improved accuracy with all relative errors below 10\%, as shown in Figure \ref{Figure:HBI_KdV_xi_t} (b). The average relative estimation error is as low as 2.88\%. The comparison results verify the accuracy of model parameter estimation for tests under various conditions in multiscale system modeling.

\begin{figure}[!h]
	\centering
	\includegraphics[scale=0.8]{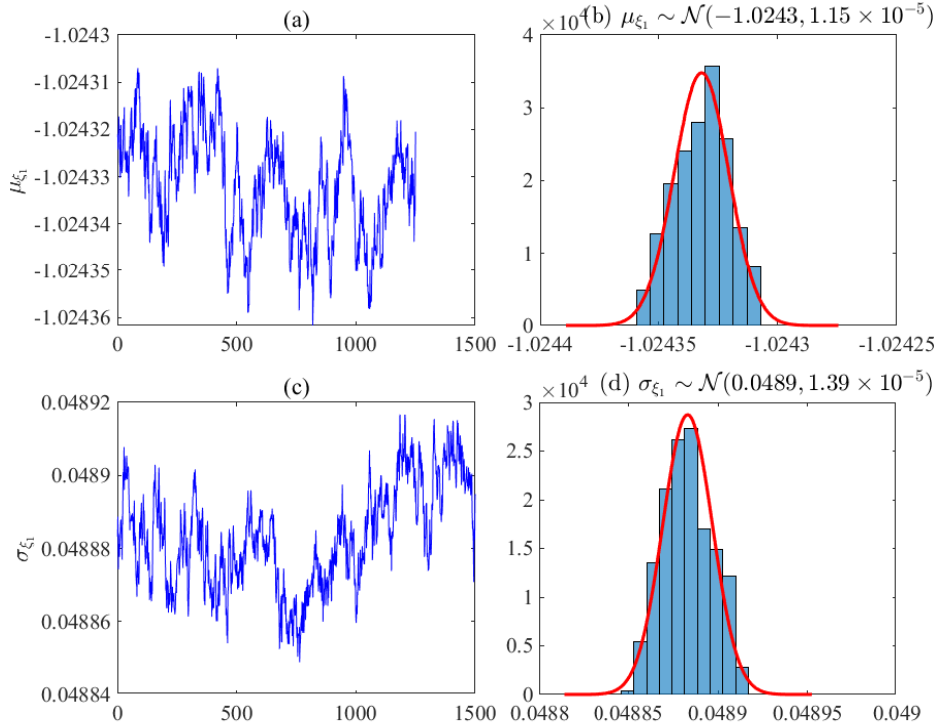}
	\caption{Samples of statistics of model parameter $\xi_1$ and their distribution. (a) samples of the mean of $\xi_1$; (b) distribution of samples in (a) and the fitted Gaussian distribution; (c) samples of the standard deviation of $\xi_1$; (d) distribution of samples in (c) and the fitted Gaussian distribution.}
	\label{Figure:HBI_KdV_mu_xi1}
\end{figure}

\begin{figure}[!h]
	\centering
	\includegraphics[scale=0.8]{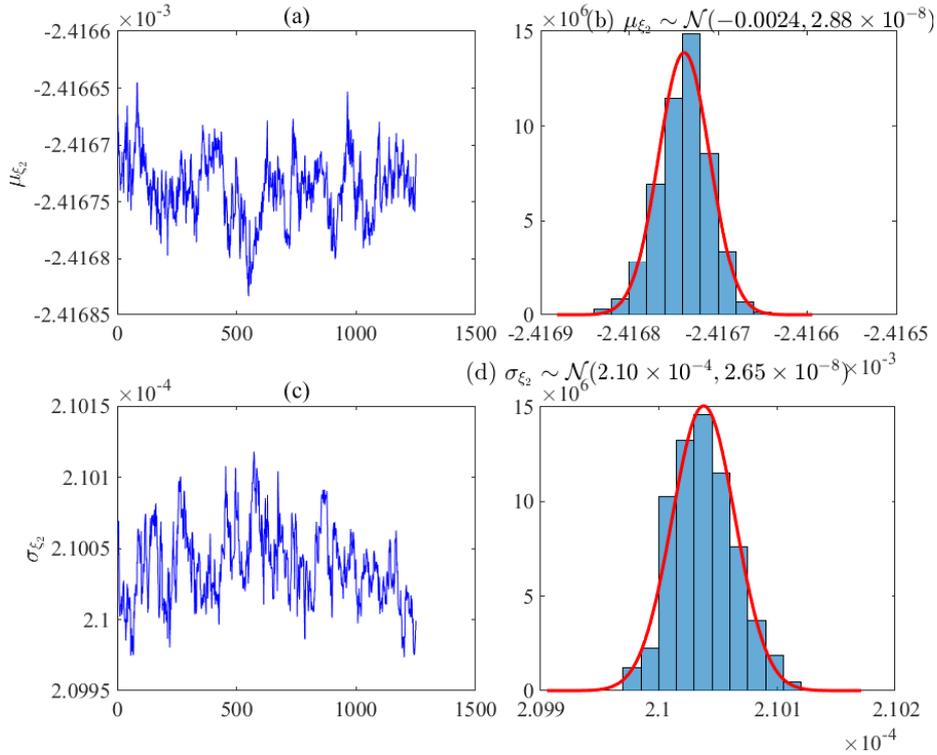}
	\caption{Samples of statistics of model parameter $\xi_2$ and their distribution. Please refer to the caption of Figure \ref{Figure:HBI_KdV_mu_xi1} for more information.}
	\label{Figure:HBI_KdV_mu_xi2}
\end{figure}

Figures \ref{Figure:HBI_KdV_mu_xi1} and \ref{Figure:HBI_KdV_mu_xi2} show the results of Bayesian inference of the statistics of model parameters $\xi_1$ and $\xi_2$, respectively. Figures \ref{Figure:HBI_KdV_mu_xi1} (b) and (d) show that the model parameter $\xi_1$ has an estimated mean of -1.0243 and an estimated standard deviation of 0.0489, which are very close to the reference values, i.e., -1.00 and 0.05, respectively. An accurate estimation of statistics of the model parameter $\xi_2$ can also be observed in Figure \ref{Figure:HBI_KdV_mu_xi2}. The accurate estimation of statistics of model parameters approve that the proposed HBI method for multiscale system modeling can efficiently evaluate the state of system governing model including its uncertainty under varying conditions, especially with large and various measurement noise levels. 

\section{Summary and Further Discussions} \label{Section:Conclusion}
In this study, a Parsimony-Enhanced Sparse Bayesian Learning method (i.e., the \textsc{PeSBL} method) is proposed for robust data-driven discovery of governing PDEs. The  \textsc{PeSBL} method is advantageous over most existing methods in discovering the accurate governing models of new complex dynamical systems. Compared with deterministic PDE learning methods, this SBL-based method automatically promotes sparsity via specifying an independent prior for each model parameter and automatically pruning redundant parameters in sequential iterations. However, promoting sparsity alone is not necessarily sufficient for discovering an accurate governing model for a complex dynamical system. Therefore, based on existing SBL-based PDE learning methods in the literature, the \textsc{PeSBL} method further enhances parsimony via modifying the evaluation criteria in the sequential solving algorithm. This modification accounts for the complexity of each candidate model term and thus prevents adding complex terms for marginal increase of regression accuracy in each iteration. Thus, this method fundamentally avoids the model pruning procedure in most existing deterministic/stochastic PDE learning methods that inevitably falls into the lemma of hyperparameter tuning. Additionally, advanced signal preprocessing techniques and Bayesian model updating (BMU) further increase the robustness of PDE learning results. Results of numerical simulations show that the accurate parsimonious governing PDEs can be correctly identified from noisy data for several canonical dynamical systems using the proposed \textsc{PeSBL} method.

Moreover, the stochastic identification scheme in the \textsc{PeSBL} method enables quantifying model uncertainties, propagating uncertainties to system prediction, and conducting probabilistic system diagnosis and prognosis. In reality, the investigated system may vary with the change of exterior conditions such as temperature. As a result, the system model parameters may be intrinsically stochastic. In this study, the  intrinsic system model uncertainty is simulated via multiscale Bayesian system modeling through Hierarchical Bayesian Inference (HBI). A numerical case study is conducted with the traveling wave system characterized by the KdV equation. The results show that the system uncertainty under varying conditions can be accurately evaluated even with large measurement noise using the established multiscale Bayesian system modeling framework. 

The future work may include applying the  \textsc{PeSBL} method in real operating dynamical systems to further examine or improve its effectiveness in knowledge discovery.

\section{Acknowledgments}
The research reported in this paper was supported by funds from NASA University Leadership Initiative Program (Contract No. NNX17AJ86A, Project Officer: Dr. Anupa Bajwa, Principal Investigator: Dr. Yongming Liu). The support is gratefully acknowledged.

\bibliographystyle{elsarticle-num}
\bibliography{PE-SBL_PDE_Learn}

\begin{thebibliography}{10}
\expandafter\ifx\csname url\endcsname\relax
  \def\url#1{\texttt{#1}}\fi
\expandafter\ifx\csname urlprefix\endcsname\relax\def\urlprefix{URL }\fi
\expandafter\ifx\csname href\endcsname\relax
  \def\href#1#2{#2} \def\path#1{#1}\fi

\bibitem{rudy2017data}
S.~H. Rudy, S.~L. Brunton, J.~L. Proctor, J.~N. Kutz, Data-driven discovery of
  partial differential equations, Science Advances 3~(4) (2017) e1602614.

\bibitem{long2019pde}
Z.~Long, Y.~Lu, B.~Dong, Pde-net 2.0: Learning pdes from data with a
  numeric-symbolic hybrid deep network, Journal of Computational Physics 399
  (2019) 108925.

\bibitem{schmidt2009distilling}
M.~Schmidt, H.~Lipson, Distilling free-form natural laws from experimental
  data, science 324~(5923) (2009) 81--85.

\bibitem{raissi2018hidden}
M.~Raissi, G.~E. Karniadakis, Hidden physics models: Machine learning of
  nonlinear partial differential equations, Journal of Computational Physics
  357 (2018) 125--141.

\bibitem{maslyaev1903data}
M.~Maslyaev, A.~Hvatov, A.~Kalyuzhnaya, Data-driven pde discovery with
  evolutionary approach.(2019), arXiv preprint arXiv:1903.08011.

\bibitem{atkinson2019data}
S.~Atkinson, W.~Subber, L.~Wang, G.~Khan, P.~Hawi, R.~Ghanem, Data-driven
  discovery of free-form governing differential equations, arXiv preprint
  arXiv:1910.05117.

\bibitem{hasan2020learning}
A.~Hasan, J.~M. Pereira, R.~Ravier, S.~Farsiu, V.~Tarokh, Learning partial
  differential equations from data using neural networks, in: ICASSP 2020-2020
  IEEE International Conference on Acoustics, Speech and Signal Processing
  (ICASSP), IEEE, 2020, pp. 3962--3966.

\bibitem{xu2020dlga}
H.~Xu, H.~Chang, D.~Zhang, Dlga-pde: Discovery of pdes with incomplete
  candidate library via combination of deep learning and genetic algorithm,
  Journal of Computational Physics (2020) 109584.

\bibitem{rudy2019data}
S.~Rudy, A.~Alla, S.~L. Brunton, J.~N. Kutz, Data-driven identification of
  parametric partial differential equations, SIAM Journal on Applied Dynamical
  Systems 18~(2) (2019) 643--660.

\bibitem{chen2020deep}
Z.~Chen, Y.~Liu, H.~Sun, Deep learning of physical laws from scarce data, arXiv
  preprint arXiv:2005.03448.

\bibitem{bekar2021peridynamics}
A.~C. Bekar, E.~Madenci, Peridynamics enabled learning partial differential
  equations, Journal of Computational Physics (2021) 110193.

\bibitem{berg2019data}
J.~Berg, K.~Nystr{\"o}m, Data-driven discovery of pdes in complex datasets,
  Journal of Computational Physics 384 (2019) 239--252.

\bibitem{xiong2019data}
B.~Xiong, H.~Fu, F.~Xu, Y.~Jin, Data-driven discovery of partial differential
  equations for multiple-physics electromagnetic problem, arXiv preprint
  arXiv:1910.13531.

\bibitem{both2021deepmod}
G.-J. Both, S.~Choudhury, P.~Sens, R.~Kusters, Deepmod: Deep learning for model
  discovery in noisy data, Journal of Computational Physics 428 (2021) 109985.

\bibitem{zhang2021robust}
Z.~Zhang, Y.~Liu, Robust data-driven discovery of partial differential
  equations under uncertainties, arXiv preprint arXiv:2102.06504.

\bibitem{zhang2018robust}
S.~Zhang, G.~Lin, Robust data-driven discovery of governing physical laws with
  error bars, Proceedings of the Royal Society A: Mathematical, Physical and
  Engineering Sciences 474~(2217) (2018) 20180305.

\bibitem{chen2021robust}
A.~Chen, G.~Lin, Robust data-driven discovery of partial differential equations
  with time-dependent coefficients, arXiv preprint arXiv:2102.01432.

\bibitem{fuentes2021equation}
R.~Fuentes, R.~Nayek, P.~Gardner, N.~Dervilis, T.~Rogers, K.~Worden, E.~Cross,
  Equation discovery for nonlinear dynamical systems: A bayesian viewpoint,
  Mechanical Systems and Signal Processing 154 (2021) 107528.

\bibitem{nayek2020spike}
R.~Nayek, R.~Fuentes, K.~Worden, E.~J. Cross, On spike-and-slab priors for
  bayesian equation discovery of nonlinear dynamical systems via sparse linear
  regression, arXiv preprint arXiv:2012.01937.

\bibitem{yuan2019machine}
Y.~Yuan, J.~Li, L.~Li, F.~Jiang, X.~Tang, F.~Zhang, S.~Liu, J.~Goncalves, H.~U.
  Voss, X.~Li, et~al., Machine discovery of partial differential equations from
  spatiotemporal data, arXiv preprint arXiv:1909.06730.

\bibitem{zhang2019robust}
S.~Zhang, G.~Lin, Robust subsampling-based sparse bayesian inference to tackle
  four challenges (large noise, outliers, data integration, and extrapolation)
  in the discovery of physical laws from data, arXiv preprint arXiv:1907.07788
  (2019) 80309--0526.

\bibitem{chen2020gaussian}
J.~Chen, L.~Kang, G.~Lin, Gaussian process assisted active learning of physical
  laws, Technometrics (2020) 1--14.

\bibitem{Bhouri2021gaussian}
M.~A. Bhouri, P.~Perdikaris, Gaussian processes meet neuralodes: a bayesian
  framework for learning the dynamics of partially observed systems from scarce
  and noisy data, arXiv preprint arXiv:2103.03385.

\bibitem{jefferys1992ockham}
W.~H. Jefferys, J.~O. Berger, Ockham's razor and bayesian analysis, American
  Scientist 80~(1) (1992) 64--72.

\bibitem{tipping2003fast}
M.~TIPPING, Fast marginal likelihood maximisation for sparse bayesian models,
  in: Proc. Ninth International Workshop on Artificial Intelligence and
  Statistics, 2003, 2003, pp. 3--6.

\bibitem{wipf2004sparse}
D.~P. Wipf, B.~D. Rao, Sparse bayesian learning for basis selection, IEEE
  Transactions on Signal processing 52~(8) (2004) 2153--2164.

\bibitem{lagergren2020learning}
J.~H. Lagergren, J.~T. Nardini, G.~Michael~Lavigne, E.~M. Rutter, K.~B. Flores,
  Learning partial differential equations for biological transport models from
  noisy spatio-temporal data, Proceedings of the Royal Society A 476~(2234)
  (2020) 20190800.

\bibitem{zanna2020data}
L.~Zanna, T.~Bolton, Data-driven equation discovery of ocean mesoscale
  closures, Geophysical Research Letters 47~(17) (2020) e2020GL088376.

\bibitem{xu2019dl}
H.~Xu, H.~Chang, D.~Zhang, Dl-pde: Deep-learning based data-driven discovery of
  partial differential equations from discrete and noisy data, arXiv preprint
  arXiv:1908.04463.

\bibitem{cao2020machine}
W.~Cao, W.~Zhang, Machine learning of partial differential equations from noise
  data, arXiv preprint arXiv:2010.06507.

\bibitem{von2011statistical}
U.~Von~Luxburg, B.~Sch{\"o}lkopf, Statistical learning theory: Models,
  concepts, and results, in: Handbook of the History of Logic, Vol.~10,
  Elsevier, 2011, pp. 651--706.

\bibitem{wagenmakers2004aic}
E.-J. Wagenmakers, S.~Farrell, Aic model selection using akaike weights,
  Psychonomic bulletin \& review 11~(1) (2004) 192--196.

\bibitem{rubinstein2016simulation}
R.~Y. Rubinstein, D.~P. Kroese, Simulation and the Monte Carlo method, Vol.~10,
  John Wiley \& Sons, 2016.

\bibitem{gilks1995markov}
W.~R. Gilks, S.~Richardson, D.~Spiegelhalter, Markov chain Monte Carlo in
  practice, Chapman and Hall/CRC, 1995.

\bibitem{cowles1996markov}
M.~K. Cowles, B.~P. Carlin, Markov chain monte carlo convergence diagnostics: a
  comparative review, Journal of the American Statistical Association 91~(434)
  (1996) 883--904.

\bibitem{raissi2019physics}
M.~Raissi, P.~Perdikaris, G.~E. Karniadakis, Physics-informed neural networks:
  A deep learning framework for solving forward and inverse problems involving
  nonlinear partial differential equations, Journal of Computational Physics
  378 (2019) 686--707.

\bibitem{baydin2017automatic}
A.~G. Baydin, B.~A. Pearlmutter, A.~A. Radul, J.~M. Siskind, Automatic
  differentiation in machine learning: a survey, The Journal of Machine
  Learning Research 18~(1) (2017) 5595--5637.

\bibitem{Zienkiewicz2005finite}
O.~Zienkiewicz, R.~Taylor, P.~Nithiarasu, The finite element method for fluid
  dynamics. 6th edition, Elsevier, 2005.

\end{thebibliography}

\end{document}